\documentclass[12pt]{article}
\usepackage{amsmath}
\usepackage{times}
\usepackage{graphicx}
\usepackage{color}
\usepackage[most]{tcolorbox}
\newcommand{\hltext}[1]{{\color{black}#1}}
\usepackage{multirow}
\usepackage{caption}
\usepackage[authoryear, round]{natbib}
\usepackage{rotating}
\usepackage{bbm}
\usepackage{latexsym}

\newcommand{\captionfonts}{\normalsize}

\makeatletter  
\long\def\@makecaption#1#2{%
  \vskip\abovecaptionskip
  \sbox\@tempboxa{{\captionfonts #1: #2}}%
  \ifdim \wd\@tempboxa >\hsize
    {\captionfonts #1: #2\par}
  \else
    \hbox to\hsize{\hfil\box\@tempboxa\hfil}%
  \fi
  \vskip\belowcaptionskip}
\makeatother   

\numberwithin{equation}{section}
\usepackage{amssymb, amsmath,amsfonts}
\usepackage{algorithmic}
\usepackage{algorithm}
\usepackage{array}
\usepackage[caption=false,font=normalsize,labelfont=sf,textfont=sf]{subfig}
\usepackage{textcomp}
\usepackage{stfloats}
\usepackage{url}
\usepackage{verbatim}
\usepackage{graphicx}
\usepackage{cite}

\usepackage{here}
\usepackage{bm}
\usepackage{amsthm}
\usepackage{orcidlink}

\theoremstyle{definition}

\theoremstyle{plain}
\newtheorem{thm}{Theorem}

\newtheorem{lem}{Lemma}
\newtheorem{cor}{Corollary}

\pdftrailerid{}
\begin{document}

\ \vspace{20mm}\\

{\LARGE \noindent Unsupervised Learning in Echo State Networks for Input  Reconstruction}

\ \\
{\bf \large Taiki Yamada}\\{Graduate School of Information Science and Technology, The University of Tokyo, Bunkyo, Tokyo 113-0033, Japan.}\\
{\bf \large Yuichi Katori}\\{School of Systems Information Science, Future University Hakodate, Hakodate, Hokkaido 041-0803, Japan.}\\
{\bf \large Kantaro Fujiwara}\\{Graduate School of Information Science and Technology, The University of Tokyo, Bunkyo, Tokyo 113-0033, Japan.}\\


\thispagestyle{empty}
\markboth{}{NC instructions}
\ \vspace{-0mm}\\
%
\begin{center} {\bf Abstract} \end{center}
Echo state networks (ESNs) are a class of recurrent neural networks in which only the readout layer is trainable, while the recurrent and input layers are fixed. 
This architectural constraint enables computationally efficient processing of time-series data.  
Traditionally, the readout layer in ESNs is trained using supervised learning with target outputs.  
In this study, we focus on input reconstruction (IR), where the readout layer is trained to reconstruct the input time series fed into the ESN.  
We show that IR can be achieved through unsupervised learning (UL), without access to supervised targets, provided that the ESN parameters are known a priori and satisfy invertibility conditions.  
This formulation allows applications relying on IR, such as dynamical system replication and noise filtering, to be reformulated within the UL framework via straightforward integration with existing algorithms.  
Our results suggest that prior knowledge of ESN parameters can reduce reliance on supervision, thereby establishing a new principle: not only by fixing part of the network parameters but also by exploiting their specific values.  
Furthermore, our UL-based algorithms for input reconstruction and related tasks are suitable for autonomous processing, offering insights into how analogous computational mechanisms might operate in the brain in principle.  
These findings contribute to a deeper understanding of the mathematical foundations of ESNs and their relevance to models in computational neuroscience.

\newpage
\section{Introduction}
\label{sec:int}
\begin{figure}[H]
\begin{center}
\includegraphics[width=100mm]{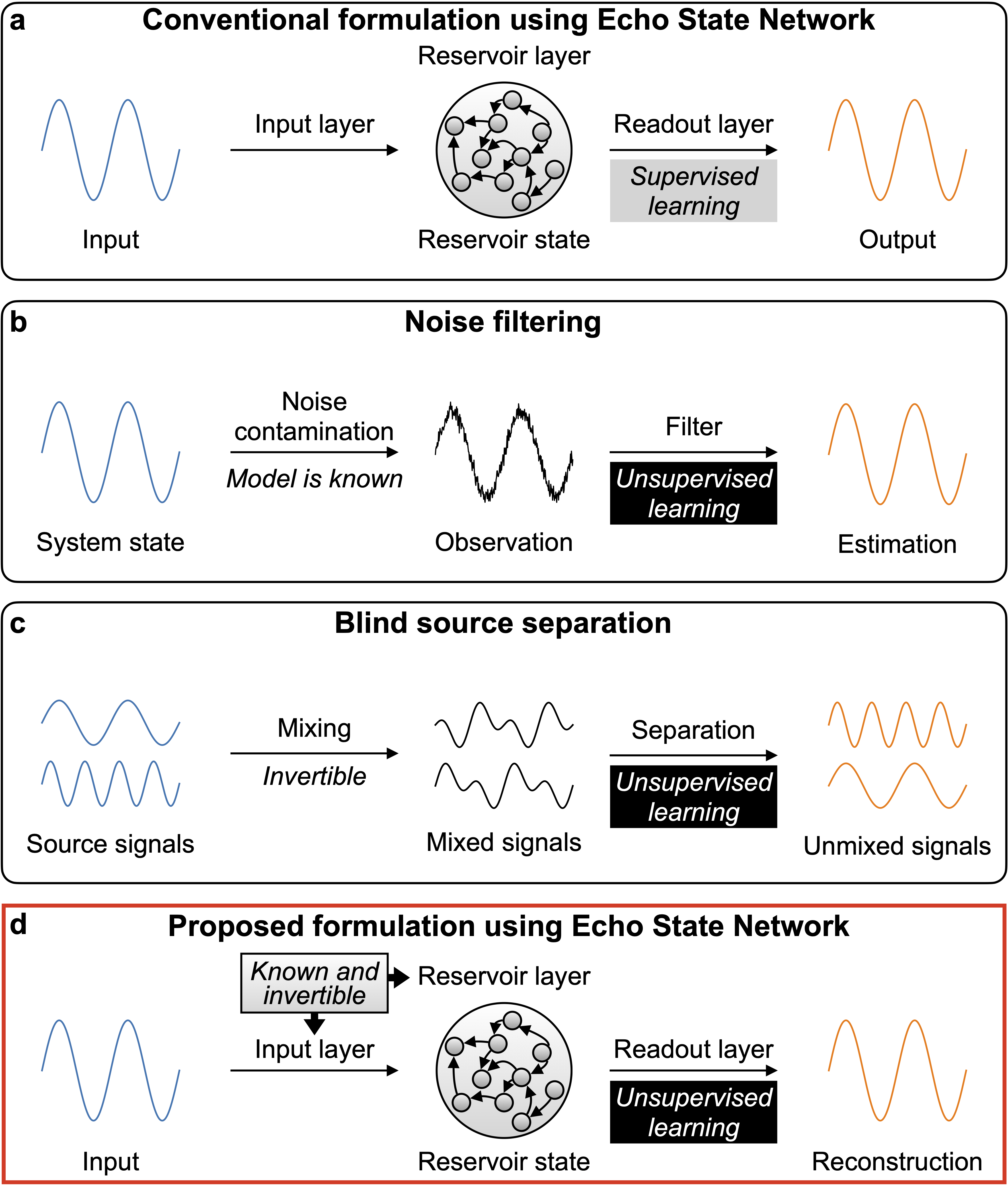}
\caption{
Overview of the proposed input reconstruction formulation in ESNs.
\textbf{a}. In typical reservoir computing frameworks, such as ESNs, training the readout layer requires supervised data of the desired output time series. Consequently, input reconstruction in ESNs has traditionally been formulated as a supervised learning problem.
\textbf{b}. In noise filtering, system states are reconstructed from noisy observations by exploiting the prior knowledge of the system and observation model.
\textbf{c}. In blind source separation, source signals are reconstructed from mixed signals assuming that the mixing process is invertible.
\textbf{d}. In this study, we propose a novel unsupervised formulation of input reconstruction in ESNs. This approach relies on two key assumptions: invertibility and availability of an ESN model for learning.
}
\label{fig:main-argument}
\end{center}
\end{figure}

Reservoir computing (RC) is an efficient computational framework for processing time-series data.  
Models such as the echo state network (ESN) \citep{jaeger2001echo} and the liquid state machine \citep{maass-real-time-2002} share a common architecture: fixed input and reservoir layers, and a trainable readout layer \citep[see, e.g.,][for reviews]{tanaka_recent_2019, nakajima2020physical}; see also Figure~\ref{fig:main-argument}\textbf{(a)}.
In standard RC settings, the readout layer is trained to reproduce a target time series, which must be defined based on the task \citep[see, e.g.,][]{schrauwen_overview_2007, lukosevicius_reservoir_2012, zhang_survey_2023, yan_emerging_2024}.  
In certain cases \citep[e.g.,][]{lu-attractor-2018, vlachas2020backpropagation, bollt2021explaining}, the input time series itself serves as the target.  
In this paper, we refer to such tasks as input reconstruction.

Input reconstruction (IR) is one of the simplest yet most fundamental tasks in RC.
The RC performance is based on two time-series processing characteristics: preservability and transformability of the input time series within RC \citep{maass-real-time-2002, jaeger:techreport2002,dambre2012information}.
IR, by definition, requires preserving the input without transformation; consequently, it is the most basic time-series processing task in RC.
Despite its simplicity, IR in RC is sufficient for learning a dynamical system that generates an input time series \citep{lu_reservoir_2017, lu-attractor-2018, rohm2021model, grigoryeva-learning-2023}.
Here, we focus on IR to investigate the fundamental properties of RC.

IR has been studied not only within the context of RC but also through two major unsupervised learning approaches: noise filtering \citep{candy2005model} and blind source separation \citep{ghahramani2003unsupervised}.
Noise filtering methods, such as the Kalman filter \citep{kalman-1960} and its nonlinear extensions \citep[see][for a review]{daum2005nonlinearkalman}, rely on prior knowledge about a system model governing the input time series.
In noise filtering, this prior knowledge is used to reconstruct the original input time series or system state using noisy observations (Figure \ref{fig:main-argument}(\textbf{b})).
In contrast, blind source separation requires no explicit system model and instead exploits statistical properties of the signals.
Examples include principal component analysis \citep{pearson1901liii}, independent component analysis \citep{jutten-blind-1991}, and slow feature analysis \citep{wiskott-slow-2002}.
Compared to noise filtering, blind source separation typically provides approximate IR by recovering the permuted and scaled versions of the original inputs.
This method relies on minimal assumptions, such as invertibility of mixing transformations, and aims to infer unobserved source signals from their observed mixtures (Figure \ref{fig:main-argument}(\textbf{c})).

Despite the conceptual relevance of IR to unsupervised learning (UL), the standard approach in RC remains supervised learning (SL) of the readout layer (Figure~\ref{fig:main-argument}\textbf{(a)}).  
Several studies have explored combining RC with noise filtering \citep[e.g.,][]{han-nonlinear-2009, tsai-robust-2010, tomizawa-combining-2021, goswami-data-driven-2021} or blind source separation \citep[e.g.,][]{antonelo-unsupervised-2009, dinh-language-2020, steiner-exploring-2023}.  
However, in most of these approaches, the unobserved original input time series is still used as supervised data for training the readout layer.
Although the above list is not exhaustive, this reliance on supervised signals remains widespread.
From the perspective of unsupervised IR, the prevailing problem formulation in RC seems conceptually inconsistent with the foundational goals of noise filtering and blind source separation, which do not assume supervision for IR.
\\

The objective of this study is to refine and advance the problem formulation of IR in RC.  
We propose a novel formulation of IR based on UL.  
Specifically, we aim to replace the conventional assumption of access to the unobserved original input time series with prior knowledge of RC parameters that satisfy certain conditions (Figure~\ref{fig:main-argument}\textbf{(d)}).
This formulation enables IR to be defined independently of any input time series, thereby serving as a standard formulation of IR within the RC framework.
To illustrate the utility of this perspective, we develop algorithms for dynamical system replication and noise filtering that do not explicitly rely on the original input time series.  
In addition, we conduct numerical experiments to demonstrate the feasibility and validity of the proposed approach.

This paper is organized as follows.
In Section~\ref{sec:uns}, we introduce ESN as the theoretical model of RC, define the IR task, and propose its reformulation within an UL framework.
In Section \ref{sec:apps}, we demonstrate that dynamical system replication and noise filtering in ESNs can be theoretically achieved via IR, thus reformulated as UL tasks.
Additionally, we provide numerical examples illustrating both tasks solved using the proposed method.
In Section \ref{sec:dis}, we discuss the implications of our findings, highlighting how IR, dynamical system replication, and noise filtering in ESNs are fundamentally redefined under the UL framework.
Finally, we present our conclusions in Section \ref{sec:con}.

\newcommand{\N}{\mathbb{N}}
\newcommand{\R}{\mathbb{R}}
\newcommand{\Z}{\mathbb{Z}}
\newcommand{\E}{\mathbb{E}}

\newcommand{\nin}{n_{\text{in}}}
\newcommand{\nr}{n_{\text{r}}}
\newcommand{\nout}{n_{\text{out}}}

\newcommand{\mcR}{\mathcal{R}}
\newcommand{\mcD}{\mathcal{D}}
\newcommand{\mcO}{\mathcal{O}}
\section{Unsupervised Input Reconstruction}  
\label{sec:uns}  

In this section, we formalize UL for IR in RC.  
Initially, we introduce ESN \citep{jaeger2001echo} as a numerically implementable and mathematically tractable RC model.  
Next, we formalize IR in ESNs.  
Finally, we characterize the mathematical conditions under which UL becomes feasible for IR in ESNs.

Here, we describe the notation used throughout this paper.  
We denote the (vector) value of variable $x$ at time $t \in \Z$ as $x_t$; we denote the horizontal concatenation of $T$ column vectors $x_t, \dots, x_{t+T-1} \in \R^n$ as $\mathcal{X}_{t, t+T-1} \in \R^{n \times T}$, using the capital calligraphic letter corresponding to $x$.  
Let $f$ be a map.
We use the notation $f(\mathcal{X}_{t, t+T-1})$ to represent the horizontal concatenation of vectors $f(x_t), \dots, f(x_{t+T-1})$.  

\subsection{Echo State Network}  
ESN is a recurrent neural network sequentially driven by an input time series.  
The ESN input at time $t$ is a vector denoted as $d_t \in \R^{\nin}$.  
The state of the ESN (reservoir state) at time $t$ is also a vector denoted as $r_t \in \R^{\nr}$.  
The reservoir states are sequentially determined using the following equation:  
\begin{alignat}{1}
    \label{ESN}
    r_{t+1} = g(d_t, r_t) := \sigma\left(Ad_t + Br_t\right),
\end{alignat}  
where $\sigma: \R^{\nr} \to \R^{\nr}$, $A \in \R^{\nr \times \nin}$, and $B \in \R^{\nr \times \nr}$.  
The activation function $\sigma$ and the weight matrices $A$ and $B$ are fixed for each ESN; they are referred to as the activation, input, and reservoir layers, respectively.

The ESN output at time $t$ is a vector defined as $W r_t \in \R^{\nout}$, where $W \in \R^{\nout \times \nr}$.  
Matrix $W$ is learnable, which is referred to as the readout layer.  
Let $o_t \in \R^{\nout}$ denote the desired ESN output at time $t$.  
$W$ is learned so that the ESN output approximates the desired output: $\|W r_t - o_t\|^2 \approx 0$, where $\|\cdot\|$ is the Euclidean norm of a vector.

The standard approach to learning $W$ is the least squares method, which requires the desired output data $o_t$.
The least squares solution for the readout layer parameter that minimizes $\|W r_t - o_t\|^2$ is
\begin{alignat}{1}
    \label{eq:st-readout}
    W_{\mcO} := \mcO_{1,T} \mcR_{1,T}^+ \in \R^{\nout \times \nr},
\end{alignat}
where $\mcR_{1,T}^+$ is the Moore--Penrose inverse of $\mcR_{1,T}$.  
Notation $W_{\mcO}$ emphasizes that it uses samples of desired outputs $\mcO_{1,T}$.

\subsection{Input Reconstruction}
In IR, we consider the case where the desired ESN output is equal to its input, i.e., $o_t = d_t\;(\nout = \nin)$ for all time steps $t$.  
Then, using the standard approach introduced in Equation \ref{eq:st-readout}, the solution for IR is
\begin{alignat}{1}
    \label{eq:input-rec-readout}
    W_{\mcD} := \mcD_{1,T} \mcR_{1,T}^+ \in \R^{\nout \times \nr}.
\end{alignat}
Notation $W_{\mcD}$ emphasizes that its calculation requires samples obtained from the original input time series $\mcD_{1,T}$.  
Thus, the standard approach for IR in ESN assumes that samples obtained from the original input time series $\mcD_{1,T}$ are available during the training of the readout layer.  
Nevertheless, the objective remains to reconstruct the original input during testing.
We address the distinction in the problem settings between training and testing by introducing a training method for the readout layer that does not explicitly rely on the original inputs $\mcD_{1,T}$.

\subsection{Unsupervised Learning for Input Reconstruction}
\label{subsec:IR}

The most vital key point of our method is that the original input $d_t$ can be expressed using reservoir state $r_t$ under certain invertibility assumptions. Specifically, we have the following lemma:

\begin{lem}
    \label{key-lem}
    $d_t = A^+\left[\sigma^{-1}(r_{t+1}) - Br_t\right]$ for all $t \in \Z$ if $\sigma$ is invertible and $\text{rank}(A) = \nin$.
\end{lem}

\begin{proof}
    Solving the ESN state update Equation \ref{ESN} $r_{t+1} = \sigma\left(Ad_t + Br_t\right)$ for $d_t$ yields:
    \begin{alignat*}{2}
        r_{t+1} &= \sigma\left(Ad_t + Br_t\right)&\\
        \sigma^{-1}(r_{t+1}) &= Ad_t + Br_t&(\because \sigma\text{ is invertible})&\\
        \sigma^{-1}(r_{t+1}) - Br_t &= Ad_t&\\
        A^+[\sigma^{-1}(r_{t+1}) - Br_t] &= d_t&(\because A\text{ has full column rank})
    \end{alignat*}
\end{proof}

Lemma \ref{key-lem} implies the following corollary using concatenated notations:
\begin{cor}
    \label{key-lem-concat}
    $\mcD_{1,T} = A^+\left[\sigma^{-1}(\mcR_{2,T+1}) - B\mcR_{1,T}\right]$ for all $t \in \Z$ if $\sigma$ is invertible and $\text{rank}(A) = \nin$.
\end{cor}

Then, we obtain the following theorem, which describes UL for IR:
\begin{thm}
\label{key-thm}
Assume that the activation function \( \sigma \) is invertible, and \( \operatorname{rank}(A) = \nin \).  
Then, the least-squares solution for IR in ESNs is given by
\begin{align}
\label{eq:Wr-1}
W_{\mcR} = A^+\left[\sigma^{-1}(\mcR_{2,T+1}) - B\mcR_{1,T}\right] \mcR_{1,T}^+.
\end{align}
\end{thm}
\begin{proof}
    From Corollary~\ref{key-lem-concat}, we have the identity
    $
    \mcD_{1,T} = A^+\left[\sigma^{-1}(\mcR_{2,T+1}) - B\mcR_{1,T}\right],
    $
    which expresses the input time series explicitly in terms of the ESN parameters and internal states.
    Substituting this expression into the least-squares solution for input reconstruction \ref{eq:input-rec-readout},
    $
    W_{\mcD} = \mcD_{1,T} \mcR_{1,T}^+,
    $
    yields the desired formulation of the readout matrix \( W_{\mcR} \) for unsupervised input reconstruction.
\end{proof}

The additional regularity condition for matrix $\mcR_{1,T}$ yields an alternative form of Theorem \ref{key-thm}, stated below.

\begin{thm}
\label{key-thm-cor}
Assume that the activation function \( \sigma \) is invertible, \( \operatorname{rank}(A) = \nin \), and \( \operatorname{rank}(\mcR_{1,T}) = \nr \).  
Then, the least-squares solution for IR in ESNs is given by
\begin{align}
\label{eq:Wr}
W_{\mcR} = A^+\left[\sigma^{-1}(\mcR_{2,T+1})\mcR_{1,T}^+ - B\right].
\end{align}
\end{thm}

\begin{proof}
Since \( \operatorname{rank}(\mcR_{1,T}) = \nr \), we have the identity \( \mcR_{1,T} \mcR_{1,T}^+ = I \).  
Substituting this into the expression for \( W_{\mcR} \) in Theorem~\ref{key-thm} yields Equation \ref{eq:Wr}.
\end{proof}

Theorems \ref{key-thm} and \ref{key-thm-cor} show that the readout layer can be trained to solve IR without explicitly relying on the original input $d_t$; instead, the fixed ESN properties $\sigma$, $A$, and $B$ can be used.  
Hence, IR in ESNs can be formalized as UL, not SL.

\hltext{
Our theory of UL in ESN for IR relies on three independent regularity conditions:  
(i) invertibility of the activation function $\sigma$,  
(ii) full column rank of the input weight matrix $A$, and  
(iii) full row rank of the reservoir state matrix $\mcR_{1,T}$.

Conditions (i) and (ii) are required to transform the readout matrix from $W_{\mcD}$ (Equation~\ref{eq:input-rec-readout}) to $W_{\mcR}$ (Equation~\ref{eq:Wr-1}).  
This transformation involves reconstructing the ESN input sequence $\mcD_{1,T}$, which is used as targets in supervised training, from the reservoir states $\mcR_{1,T+1}$ (see Corollary~\ref{key-lem-concat}).  
Violations of these conditions result in incomplete input recovery and degraded IR performance (Figure \ref{fig:S1-RRMSE}).

In practice, condition (ii) is generally satisfied, since random matrices drawn from distributions with positive variance are almost surely full rank.  
In contrast, condition (i) is more critical: commonly used activation functions such as ReLU are not globally invertible.  
Nevertheless, we numerically confirmed that IR remains feasible under ReLU, albeit with reduced precision. Despite the loss in accuracy, the reconstruction still performs well above chance (Figure \ref{fig:S1-hist}, \ref{fig:S1-RRMSE}), indicating meaningful reconstruction capability in non-invertible settings.  
See Supplementary Material for details.

Condition (iii) is necessary to derive the simplified form of the readout matrix (Equation~\ref{eq:Wr}) from the general expression (Equation~\ref{eq:Wr-1}).
This condition may fail when the input lacks complexity or when the reservoir states exhibit insufficient diversity due to the echo state property.  
However, we found that injecting noise into the ESN input improves the effective rank of $\mcR_{1,T}$ and mitigates this issue (Figure \ref{fig:S2-RRMSE}).
When conditions (i) and (ii) are satisfied but condition (iii) is not, the SL solution $W_{\mcD}$ and the intermediate UL solution $W_{\mcR}$ (Equation~\ref{eq:Wr-1}) remain mathematically equivalent, while $W_{\mcR}$ (Equation~\ref{eq:Wr}) deviates.  
Importantly, this deviation does not necessarily imply performance degradation.  
We observed that in such ill-conditioned cases, the UL solution $W_{\mcR}$ (Equation~\ref{eq:Wr}) outperforms its intermediate counterpart (Equation~\ref{eq:Wr-1}), which is guaranteed to match the SL solution (Figure \ref{fig:S2-RRMSE}).
This highlights the numerical advantage of the proposed UL-based IR framework.  
Additional results are provided in the Supplementary Material.
}

\subsubsection{Remarks}
Theorem \ref{key-thm-cor} reveals that the least squares solution of IR in ESNs can be written as $W_{\mcR} = A^+(\hat{B}_T - B)$, where $\hat{B}_T$ minimizes the following square loss function:
\begin{alignat}{1}
    \label{eq:loss}
    L_{\sigma, T}(\hat{B}) := \sum_{t=1}^{T-1}\| \sigma^{-1}(r_{t+1}) - \hat{B}r_t\|^2.
\end{alignat}
Note that $\hat{B}_T$ can be calculated without using original inputs $d_t$, $A$, and $B$.  
Apart from using the Moore--Penrose inverse of a matrix, numerous optimizations are available to obtain the minimizer of the loss function $L_{\sigma, T}(\hat{B})$.  
For example, the recursive least square (RLS) algorithm \citep{haykin2002adaptive} sequentially updates $(\hat{B}_t, P_t)$ to $(\hat{B}_{t+1}, P_{t+1})$ based on newly obtained data $(r_t, \sigma^{-1}(r_{t+1}))$, where $P_T$ is the recursive estimation of the precision matrix of the reservoir state.
Hence, a sequential algorithm for IR in ESNs based on UL is presented (Algorithm \ref{alg:RLS-input-rec}).

\begin{algorithm}[H]
\caption{RLS for UL-based IR in ESN}\label{alg:RLS-input-rec}
\begin{algorithmic}
\STATE 
\STATE \textbf{Input: }$r_1, \dots, r_{T}$, $\sigma^{-1}$, $A$, $B$
\STATE \textbf{Output: }$(W_{\mcR})_1, \dots, (W_{\mcR})_T$
\STATE $\hat{B}_1 = B$ and $P_1 = I$
\FOR{$t = 1, \dots, T-1$}
\STATE $(W_{\mcR})_t = A^+(\hat{B}_t - B)$
\STATE $v_t = \sigma^{-1}(r_{t+1}) - \hat{B}_tr_t$
\STATE $g_t = P_tr_t / (1 + r_t^\top P_tr_t)$
\STATE $P_{t+1} = (I - g_tr_t^\top)P_t$ 
\STATE $\hat{B}_{t+1} = \hat{B}_t + v_tg_t^\top$
\ENDFOR
\end{algorithmic}
\end{algorithm}

\subsubsection{Numerical Experiments}

We conducted numerical experiments using Algorithm \ref{alg:RLS-input-rec}.  
For ESN, we considered a 1-dimensional input $d_t$, which is defined as follows:
\begin{alignat*}{1}
    d_t = \begin{cases}
        \cos(\pi t/50) &1 \leq t < 400,\\
        \cos(\pi t/100) + \sin(\pi t/25) &400 \leq t < 800,\\
        \cos^9(\pi t/50) &800 \leq t \leq 1200.
    \end{cases}
\end{alignat*}

We used a 50-dimensional ESN with the element-wise hyperbolic tangent function ($\tanh$) as the activation function $\sigma$.  
We initialized the elements of the input layer $A$ using i.i.d. samples obtained from $\mathcal{N}(0, 0.02)$.  
Let $B_0 \in \R^{50 \times 50}$ be a matrix with elements being i.i.d. samples obtained from $\mathcal{N}(0, 1)$.  
We set $B = 0.9 \cdot \rho(B_0) \cdot B_0$, where $\rho(B_0)$ denotes the spectral radius of $B_0$.  
We conducted numerical experiments for 10 different implementations of parameters $A$ and $B$.  
We applied the RLS Algorithm \ref{alg:RLS-input-rec} to obtain the readout layer $(W_{\mcR})_t$ without explicitly using the original input $d_t$.  
Thus, $(W_{\mcR})_t r_t$ represents the ESN output for IR at time $t$.

\begin{figure}[H]
\begin{center}
\includegraphics[width=100mm]{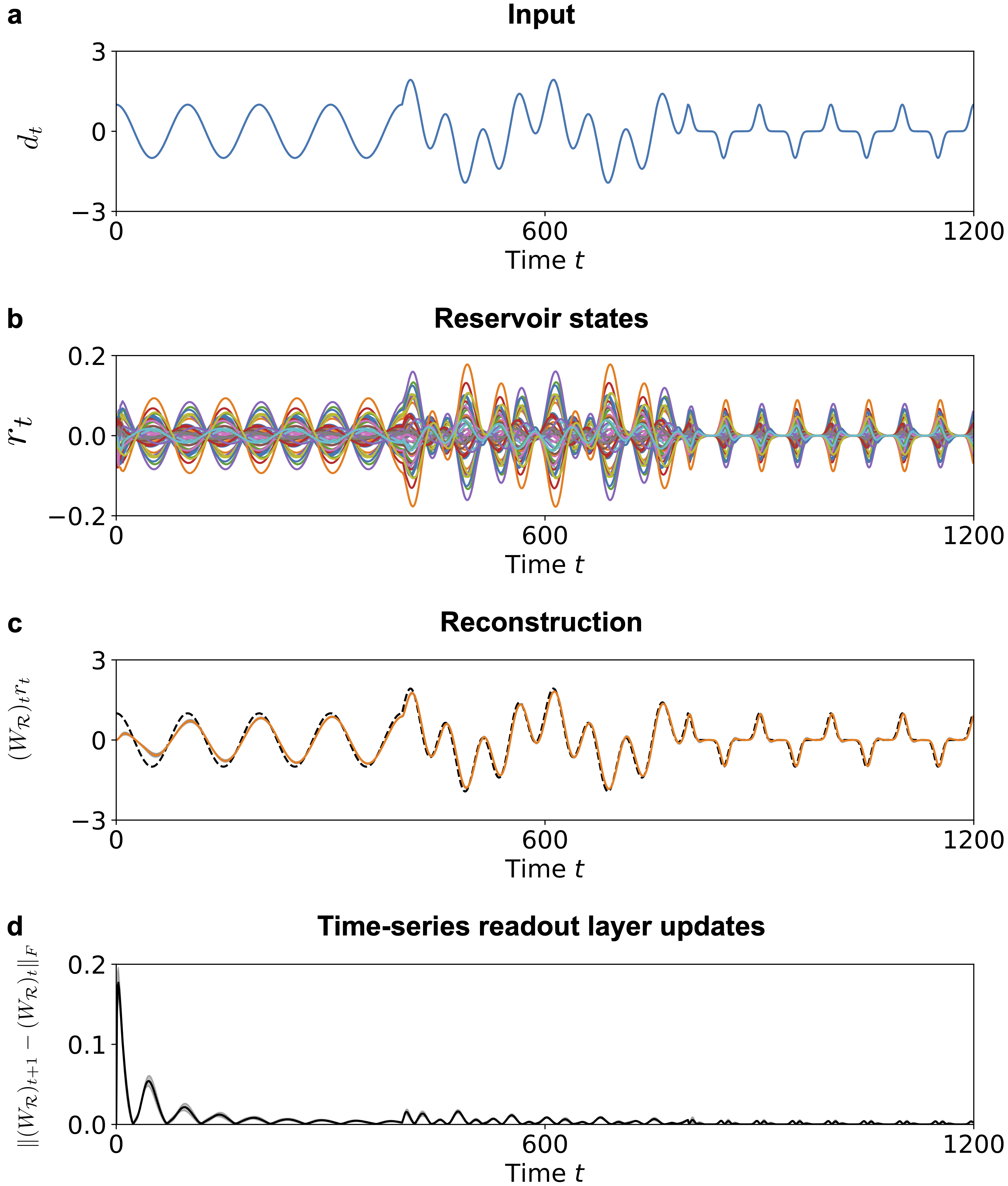}
\caption{Numerical example of Algorithm \ref{alg:RLS-input-rec} used for input reconstruction based on unsupervised learning.
    \textbf{a}. Time series $d_t$ of the ESN input. 
    \textbf{b}. Example of the reservoir state time series $r_t$ of the ESN. Each element of $r_t$ is represented in the plot with a different color.
    \textbf{c}. Average time series of the ESN output $(W_{\mcR})_t r_t$ (solid orange line) and its standard deviation across 10 trials (shaded region). The original input time series $d_t$ is plotted for reference (black dashed line).
    \textbf{d}. Average magnitude of the time-series readout layer updates $\|(W_{\mcR})_{t+1} - (W_{\mcR})_t\|_F$ (black solid line) and its standard deviation across 10 trials (shaded region).
}
\label{fig:num-input-rec}
\end{center}
\end{figure}

Figure \ref{fig:num-input-rec} shows the results obtained from Algorithm \ref{alg:RLS-input-rec} using IR based on UL.  
We used an input that changed every 400 steps (Figure \ref{fig:num-input-rec}(\textbf{a})).  
Driven by the input, the reservoir state evolved (Figure \ref{fig:num-input-rec}(\textbf{b})).  
As learning progressed, the ESN output approached the original input on average (Figure \ref{fig:num-input-rec}(\textbf{c})).  
Despite variations in the input, the readout weights exhibited only minimal changes beyond 400 steps, as averaged over trials with different ESN parameterizations (Figure \ref{fig:num-input-rec}(\textbf{d})).  
By conducting numerical experiments, we verified that IR in ESNs can be achieved without explicitly using the original input, thus validating UL.

\newpage
\section{Applications of Unsupervised Input Reconstruction}
\label{sec:apps}

In Section~\ref{sec:uns}, we formulated IR in ESNs as a UL problem.
This reformulation enables tasks that rely on IR to be addressed within the UL framework, independently of the original input sequence.
In this section, we explore two such applications: dynamical system replication and noise filtering.

We first consider the replication of deterministic dynamical systems that generate the input time series.
We show that this task can be reformulated as a UL problem.

Then, we address noise filtering, where the goal is to improve the robustness of IR when the input is corrupted by noise.
We demonstrate that the replicated system obtained via UL can be used for effective noise suppression.

\subsection{Dynamical System Replication}
\label{subsec:dyn-rep}
\subsubsection{Problem Formulations}
\begin{figure}[H]
\begin{center}
\includegraphics[width=120mm]{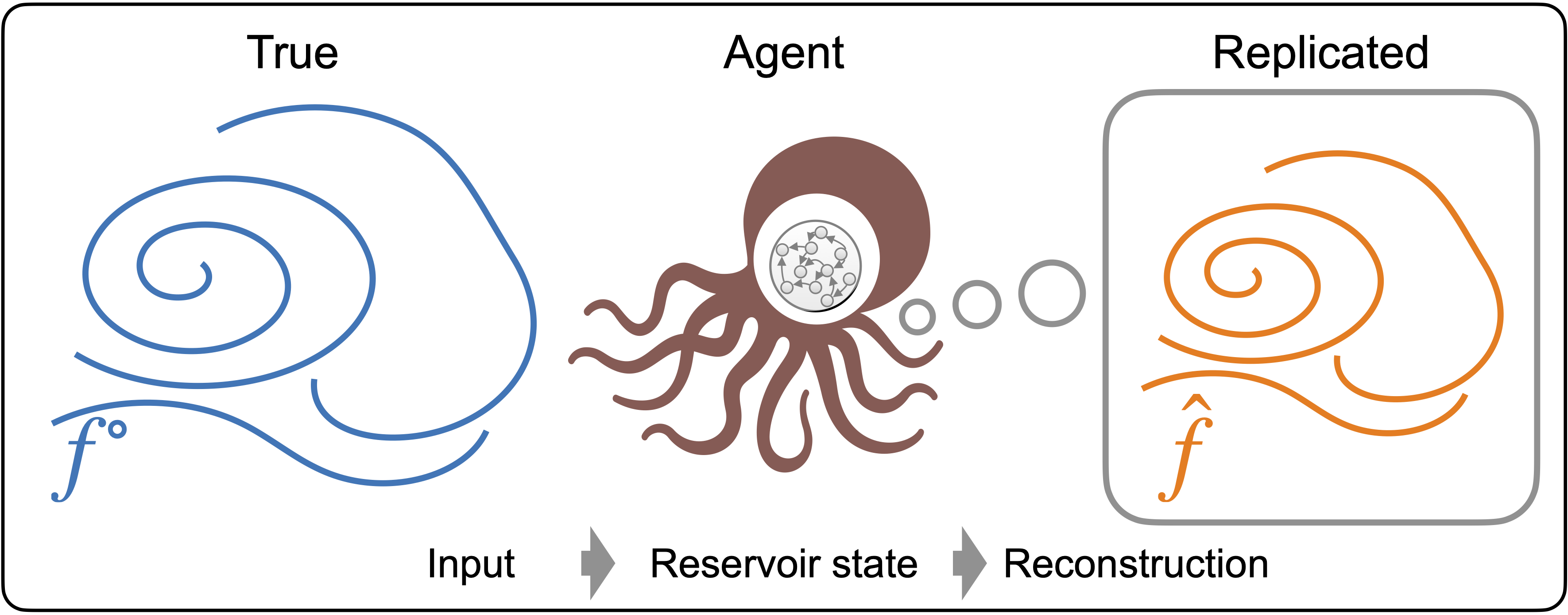}
\caption{Schematics of dynamical system replication, where an agent constructs a replicated dynamical system $\hat{f}$ using the observations of states of the true dynamical system $f^{\circ}$.
}
\label{fig:problem-dyn-rep}
\end{center}
\end{figure}

We begin by formulating the dynamical system replication task.
Let $(X, f^{\circ})$ be a dynamical system, where $X \subseteq \R^k$ and $f^{\circ}$ is a function from $X$ to itself.  
We refer to $f^\circ$ as the true dynamical system.  
Let $\{x_{t}\}_{t=1}^T$ be a finite orbit of the dynamical system $(X, f^\circ)$, such that $x_1 \in X$ and $x_{t+1} = f^\circ(x_t)$ for all $t=1, \dots, T-1$.  
The goal of dynamical system replication is to construct a dynamical system $(Y, \hat{f})$ based on a finite orbit $\{x_t\}_{t=1}^T$ (Figure~\ref{fig:problem-dyn-rep}).
We informally define that dynamical system replication is successful if $f^\circ$ and $\hat{f}$ are equivalent, where the notion of equivalence will be made precise in subsequent sections depending on the learning algorithm.

\subsubsection{Remarks}
An algorithm for dynamical system replication was proposed in \citep{pathak2017using}.  
The algorithm is based on supervised learning for IR in ESN with a finite orbit $\{x_{t}\}_{t=1}^T$, which is used as the ESN input.  
Consider an ESN $r_{t+1} = g(d_t, r_t)$, where $d_t = x_t$.  
We recall that $W_{\mcD}$ is a readout layer obtained using SL so that it reconstructs the input: $W_{\mcD}r_t \approx d_t = x_t$.  
The output of the algorithm is the dynamical system on $Y = \R^{\nr}$, which is defined as follows:
\begin{alignat}{1}
    \label{eq:rep-system-SL}
    \hat{f}_{\mcD}:r\mapsto g\left(W_{\mcD}r, r\right) = \sigma\left[\left(AW_{\mcD} + B\right)r\right].
\end{alignat}
By applying some conditions on the ESN, $f^{\circ}$ and $\hat{f}_{\mcD}$ are equivalent in the sense of the topological conjugacy of dynamical systems \citep{katok1995introduction}, supporting the correctness of the dynamical system replication algorithm \citep[see,e.g.,][for more details]{pathak-model-free-2018, lu-attractor-2018, lu-invertible-2020, hart-embedding-2020, kim-teaching-2021, hart-echo-2021, hara-learning-2022, grigoryeva-learning-2023, hart-generalised-2024}.

In Section \ref{sec:uns}, we showed that $W_{\mcD}$ can be reformulated as $W_{\mcR}$, which was defined in Equation \ref{eq:Wr}.  
Thus, the output of the dynamical system replication algorithm is reformulated as follows:
\begin{alignat}{1}
    \label{eq:rep-system}
    \hat{f}_{\mcR}:r\mapsto g(W_{\mcR}r, r) = \sigma\left[\left(AW_{\mcR} + B\right)r\right] = \sigma\left(\hat{B}r\right).
\end{alignat}
The calculation of $W_{\mcR}$ and $\hat{B}$ does not explicitly use the finite orbit $\{x_{t}\}_{t=1}^T$ of the true dynamical system.  
This indicates that, similar to IR, dynamical system replication is also possible within the ESN architecture (Figure \ref{fig:problem-dyn-rep}).

We further note that the computation of \( \hat{B} \) in Equation \ref{eq:rep-system}, which minimizes the loss function defined in Equation~\ref{eq:loss}, does not require access to the ESN parameters \( A \) and \( B \), but only the activation function \( \sigma \).  
Since computing \( W_{\mcR} \) does require knowledge of \( A \) and \( B \), this observation implies that, interestingly, dynamical system replication is a strictly simpler task than input reconstruction in terms of internal knowledge requirements.

\subsubsection{Numerical Experiments}
We conducted numerical experiments to demonstrate dynamical system replication based on UL.
Consider the Lorenz-63 system defined by
\begin{alignat*}{1}
    \begin{cases}
        \dot{\xi} = 10(\eta - \xi),\\
        \dot{\eta} = \xi(28 - \zeta) - \eta,\\
        \dot{\zeta} = \xi\eta - \frac{8}{3}\zeta.
    \end{cases}
\end{alignat*}
Let $\phi_\tau: [\xi(t), \eta(t), \zeta(t)]^\top \mapsto [\xi(t+\tau), \eta(t+\tau), \zeta(t+\tau)]^\top$ denote its time-$\tau$ flow map.
We treat the time-$\tau$ flow map $\phi_{0.02}$ as the discrete-time dynamical system $f^\circ$.
We numerically calculated a 7000-step orbit $\{x_{t} := \phi_{0.02}^{t-1}(x_1)\}_{t=1}^{7000}$ of the true dynamical system $\phi_{0.02}$ using the 4th-order Runge--Kutta method with the initial state $x_1 = [1, 1, 1]^\top$ and a time step size of $0.02$. 
We used $\{x_{t}\}_{t=1}^{5000}$ for training and $\{x_{t}\}_{t=5001}^{7000}$ for testing.
Additionally, we used a 500-dimensional ESN with the element-wise tanh function as the activation function $\sigma$.  
We set the elements of the input layer $A$ using i.i.d. samples obtained from $\mathcal{N}(0, 0.02)$.  
Let $B_0 \in \R^{500 \times 500}$ be a matrix with elements being i.i.d. samples from $\mathcal{N}(0, 1)$.  
We set $B = 1.2 \cdot \rho(B_0) \cdot B_0$, where $\rho(B_0)$ is the spectral radius of $B_0$.  
We conducted numerical experiments for 10 different implementations of parameters $A$ and $B$, and numerically calculated $W_{\mcD}$ and $W_{\mcR}$ using Equation \ref{eq:input-rec-readout} and \ref{eq:Wr}, respectively.  
Then, we numerically calculated a 2000-step orbit $\{\hat{r}_{t}\}_{t=1}^{2000}$ of $\hat{f}_{\mcD}$ defined by Equation \ref{eq:rep-system-SL} with the initial condition $\hat{r}_1 = r_{5001}$.  
Each orbit of the 500-dimensional dynamical system $\{\hat{r}_{t}\}_{t=1}^{2000}$ of $\hat{f}_{\mcD}$ was projected onto a 3-dimensional space by calculating $\{W_{\mcD}\hat{r}_{t}\}_{t=1}^{2000}$ for visualization.  
We performed the same procedure for $\hat{f}_{\mcR}$ defined by Equation \ref{eq:rep-system}. 

\begin{figure}[H]
\begin{center}
\includegraphics[width=120mm]{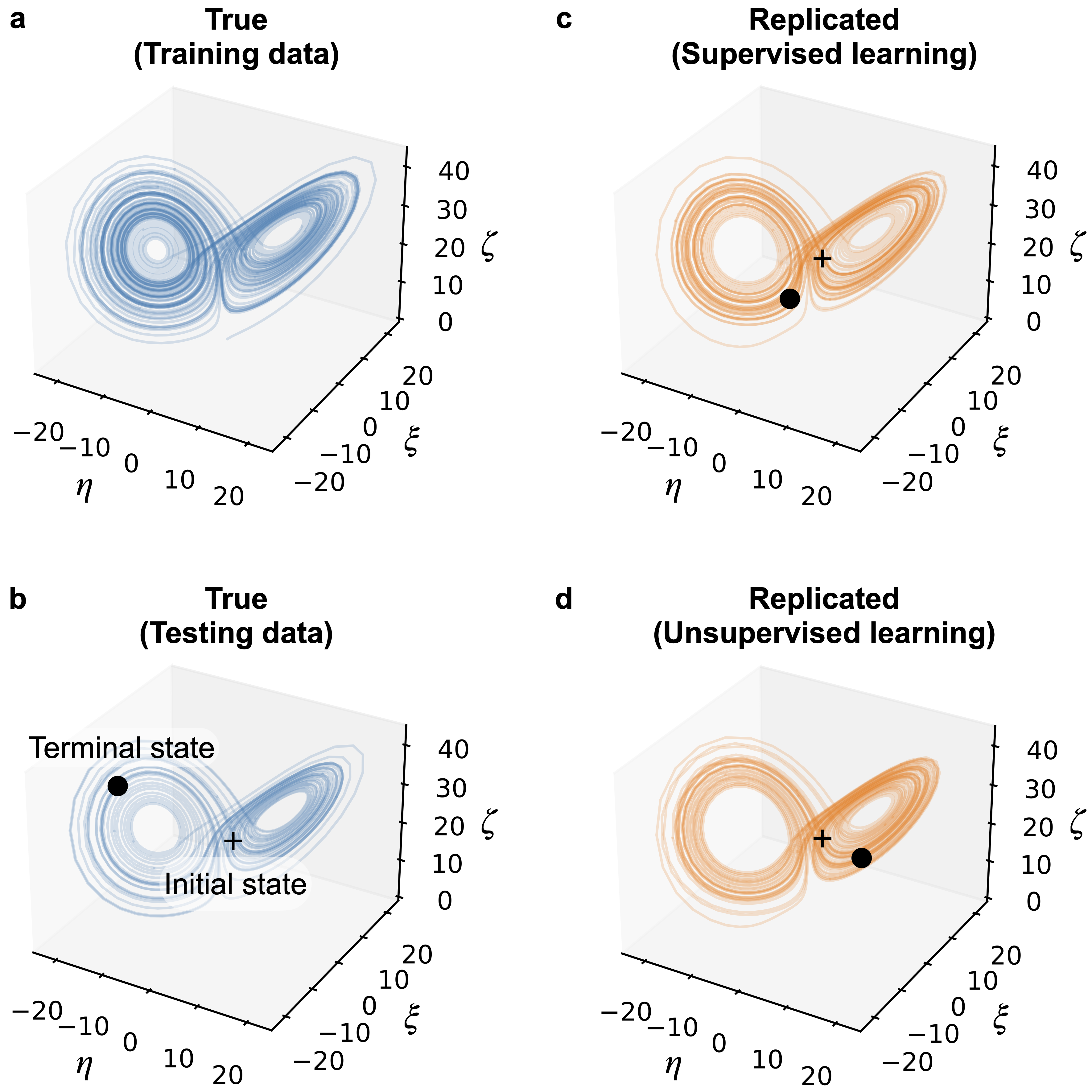}
\caption{Typical orbits of the dynamical systems.
    \textbf{a}. Orbit of the true dynamical system $f^{\circ}$ (training).
    \textbf{b}. Orbit of the true dynamical system $f^{\circ}$ (testing).
    \textbf{c}. Projected orbit of the replicated system $\hat{f}_{\mcD}$ (supervised learning).
    \textbf{d}. Projected orbit of the replicated system $\hat{f}_{\mcR}$ (unsupervised learning), with the same initial condition as in (c).
}
\label{fig:num-dyn-rep}
\end{center}
\end{figure}

We used the inverse map of $\sigma = \tanh$ and the Moore--Penrose inverse of matrix $A$ to calculate $W_{\mcR}$ using Equation \ref{eq:Wr}.  
It should be noted that, in the numerical experiments, we used numpy.arctanh and numpy.linalg.pinv employed in the Python package NumPy.  
All numerical calculations were performed with float64 precision as per the default setting of NumPy.
Figure \ref{fig:num-dyn-rep} shows a typical numerical example of the orbits of the dynamical systems.
Figures \ref{fig:num-dyn-rep}(\textbf{a}) and (\textbf{b}) show the orbits of the true dynamical system used for training and testing, respectively.
We observe that the projected orbits of both replicated systems, $\hat{f}_{\mcD}$ (Figure~\ref{fig:num-dyn-rep}\textbf{c}) and $\hat{f}_{\mcR}$ (Figure~\ref{fig:num-dyn-rep}\textbf{d}), closely resemble the true orbit (Figure~\ref{fig:num-dyn-rep}\textbf{b}) in terms of the Lorenz attractor's shape.
This observation suggests that the replication was successful in preserving the qualitative dynamics, particularly with respect to topological conjugacy.  
On the other hand, the terminal states differed across the replicated dynamical systems, even though they shared the same initial state (Figure \ref{fig:num-dyn-rep}(\textbf{c}) and (\textbf{d})).

The discrepancy between $\hat{f}_{\mcD}$ and $\hat{f}_{\mcR}$ is attributed to numerical errors in the calculation of $W_{\mcD}$ and $W_{\mcR}$ (Table~\ref{tab:dyn-rep-num-error}, 1st row).
The numerical errors stem from the approximations involved in computing the inverse tanh function (Table~\ref{tab:dyn-rep-num-error}, second row) and the Moore--Penrose inverse (third and fourth rows).

\begin{table}[H]
\caption{Numerical errors in comparing $W_{\mcD}$ and $W_{\mcR}$ across 10 trials}
\label{tab:dyn-rep-num-error}
    \centering
\begin{tabular}{lcc}
\hline
     & Min. & Max.\\\hline
     1. $\|W_{\mcR}  - W_{\mcD}\|_F$& $1.0\times 10^{-2}$ & $2.7\times 10^{-2}$\\\hline
     2. $\|\tanh(\text{arctanh}(\mcR))  - \mcR\|_F$& $3.5\times 10^{-14}$ & $3.6\times 10^{-14}$\\\hline
     3. $\|\mcR\mcR^+  - I\|_F$& $5.0\times 10^{-11}$ & $3.0\times 10^{-10}$\\\hline
     4. $\|A^+ A - I\|_F$& $3.4\times 10^{-16}$ & $1.8\times 10^{-15}$\\\hline
\end{tabular}
\end{table}


\subsection{Noise Filtering}
\label{subsec:filtering}
\subsubsection{Problem Formulations}
\begin{figure}[H]
\begin{center}
\includegraphics[width=110mm]{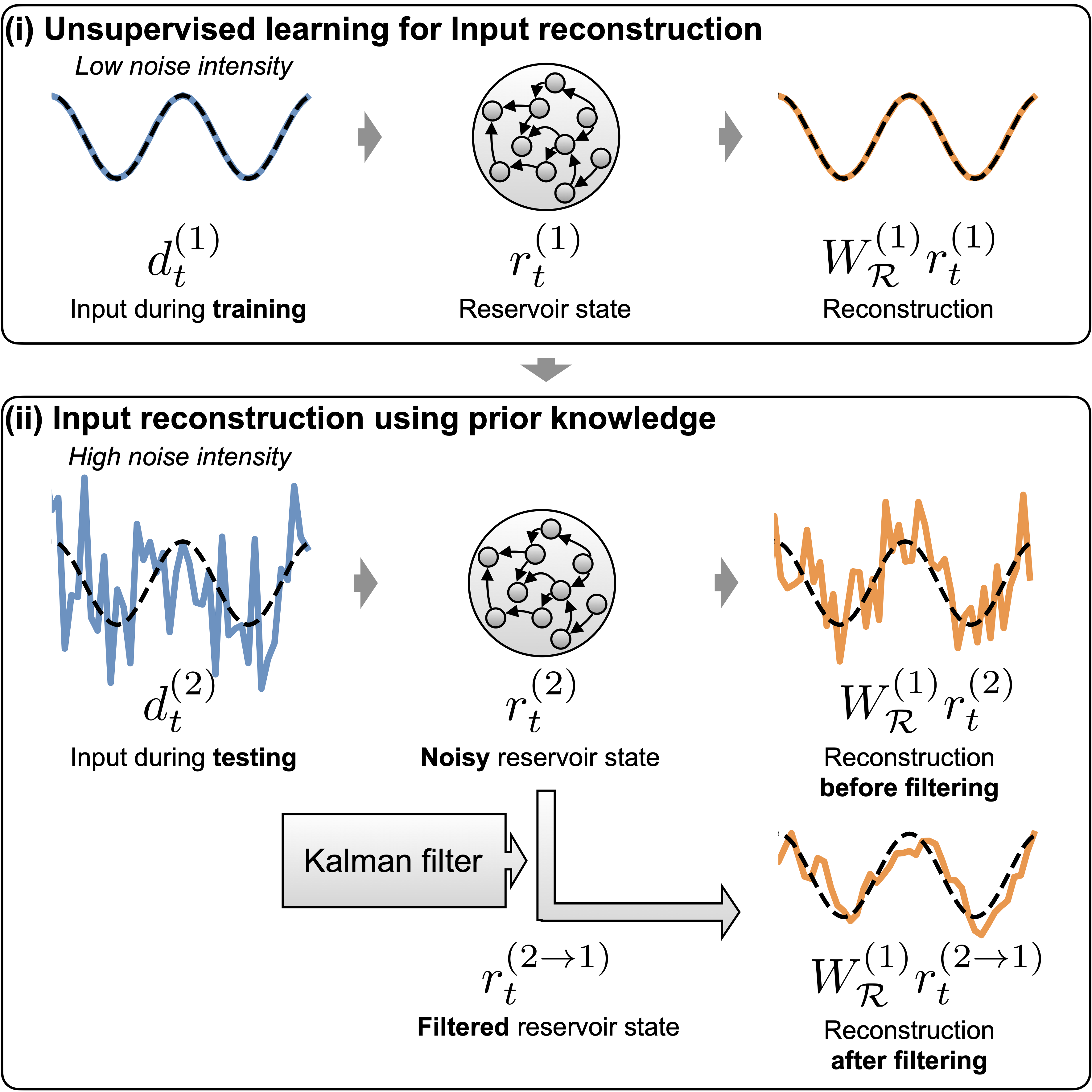}
\caption{Schematics of noise filtering in ESNs. The ESN input time series is contaminated by noise with a varying variance. The ESN reservoir states are also contaminated because of the contaminated input.
The objective of this problem is to reconstruct the noise-free input time series using the reservoir states.}
\label{fig:problem-filter}
\end{center}
\end{figure}

\newcommand{\circledtiny}[1]{\raise0.2ex\hbox{\footnotesize{\textcircled{\tiny{#1}}}}}
We proceed to formulate the noise filtering task in ESNs.
Let $d_t \in \R^{\nin}$ be the original ESN input.
Suppose that the readout layer $W_{\mcR}^{(1)}$ is trained for IR using Equation \ref{eq:Wr} with the reservoir states $r_t^{(1)}$ driven by input $d_t^{(1)}$ being contaminated by white Gaussian noise with zero mean and covariance matrix $\Sigma^{(1)} \in \R^{\nin \times \nin}$.
Consider the case where the readout layer $W_{\mcR}^{(1)}$ is retained as prior knowledge (Figure \ref{fig:problem-filter}(i)).
The purpose of noise filtering is to reconstruct the original input $d_t$ from a history of reservoir states $r_t^{(2)}$ driven by input $d_t^{(2)}$ that are contaminated by white Gaussian noise with zero mean and covariance matrix $\Sigma^{(2)}$ (Figure \ref{fig:problem-filter}(ii)).
We focus on the $\text{tr}(\Sigma^{(1)}) \leq \text{tr}(\Sigma^{(2)})$ case.

\subsubsection{Remarks}
A naive solution for noise filtering is the direct use of the readout layer $W_{\mcR}^{(1)}$ to obtain $W_{\mcR}^{(1)}r_t^{(2)}$ as a reconstruction of the input (Figure \ref{fig:problem-filter}(ii), upper row).
This results in a good reconstruction with small $\text{tr}(\Sigma^{(2)})$; however, it is not effective when $\text{tr}(\Sigma^{(2)})$ is large because $W_{\mcR}^{(1)}$ has not been trained to handle a significant amount of noise.

As shown in Section \ref{subsec:dyn-rep}, the trained readout layer $W_{\mcR}^{(1)}$ induces the autonomous dynamical system $\hat{f}_{\mcR}^{(1)}$ defined in Equation \ref{eq:rep-system}.
By definition, the dynamical system $\hat{f}_{\mcR}^{(1)}$ is regarded as the evolution law of the reservoir state virtually driven by input $d_t^{(1)}$ with a noise intensity lower than that of input $d_t^{(2)}$.
Thus, intuitively, we can use $W_{\mcR}^{(1)}$ as prior knowledge in IR.
We demonstrate that prior knowledge can be effectively leveraged via Kalman filtering to enhance the performance of IR.
For a comprehensive understanding of Kalman filter theory and algorithms, we refer the reader to \citep{kalman-1960, grewal_kalman_2001}.

\hltext{
Since the dynamical system $\hat{f}_{\mcR}^{(1)}$, used as prior knowledge for noise filtering, is a nonlinear function, 
we employed a nonlinear Kalman filter, specifically the ensemble Kalman filter (EnKF) \citep{evensen_ensemble_2003}, to estimate the filtered reservoir state $r_t^{(2\to 1)}$ from the noisy reservoir state $r_t^{(2)}$.
In EnKF, the posterior distribution of $r_t^{(2\to 1)}$ conditioned on the observations $r_0^{(2)}, \dots, r_t^{(2)}$ is approximated by an ensemble of samples $r_{t/t}^{\circledtiny{1}}, \dots, r_{t/t}^{\circledtiny{M}}$, where $M \in \mathbb{N}$ denotes the ensemble size.
}

Then, the output $W_{\mcR}^{(1)}r_t^{(2\to 1)}$ is close to $W_{\mcR}^{(1)}r_t^{(1)}$, thereby yielding a more accurate reconstruction of the original input $d_t$ than that obtained from $W_{\mcR}^{(1)}r_t^{(2)}$.
To apply \hltext{EnKF}, we define the following state-space model:
\begin{align}
    \label{eq:state-space-model}
    \begin{cases}
    r_{t+1}^{(1)} &:= \hat{f}_{\mcR}^{(1)}(r_{t}^{(1)}) + w_t,\\
    r_{t}^{(2)} &:= r_{t}^{(1)} + v_t.
    \end{cases},
\end{align}
where $w_t$ and $v_t$ are assumed to be white Gaussian noise with zero mean and covariance matrices $Q \in \R^{\nr \times \nr}$ and $R \in \R^{\nr \times \nr}$, respectively.
The Kalman filter requires the specification of the covariance matrices $Q$ and $R$.
The covariance matrix $Q$ of $w_t$ is empirically calculated using $\hat{Q} = \frac{1}{T}\sum_{t=1}^T w_t w_t^\top$.
Note that, as $\hat{f}_{\mcR}^{(1)}$ can be obtained using UL (see Section \ref{subsec:dyn-rep}), the same applies to $\hat{Q}$ using the reservoir state $r_t^{(1)}$ provided in prior learning (Algorithm \ref{alg:prior-knowledge}).

\begin{algorithm}[H]
\caption{UL for IR with accuracy estimation}\label{alg:prior-knowledge}  
\begin{algorithmic}
\STATE 
\STATE \textbf{Input: }$\mcR_{1,T}^{(1)}$, $A$, $B$, $\sigma$
\STATE \textbf{Output: }$W_{\mcR}^{(1)}$, $\hat{f}_{\mcR}^{(1)}$, $\hat{Q}$
\STATE Get $W_{\mcR}^{(1)}$ from $\mcR_{1,T}^{(1)}, A, B, \sigma$ using Equation \ref{eq:Wr}
\STATE Get $\hat{f}_{\mcR}^{(1)}$ from $W_{\mcR}^{(1)}, A, B, \sigma$ using Equation \ref{eq:rep-system}
\STATE Set $\hat{Q} = \frac{1}{T}\sum_{t=1}^T w_t w_t^\top$, where $w_t = r_{t+1} - \hat{f}_{\mcR}^{(1)}(r_{t}^{(1)})$
\end{algorithmic}
\end{algorithm}

However, the covariance matrix $R$ of $v_t$ cannot be calculated in prior learning.
This is because the empirical estimation of $R$ depends on observations of $r_t^{(2)}$.
Thus, we used the adaptive updating methods \citep{mehra_approaches_1972, mohamed_adaptive_1999, akhlaghi_adaptive_2017} to calculate an empirical estimation of the covariance matrix $R$ of $v_t$ based on the samples obtained from the noisy reservoir state $r_t^{(2)}$ provided at each time.
We denote the update rate in the estimation of the covariance matrix $R$ as $\alpha_R \in [0,1]$.
Therefore, we propose Algorithm \ref{alg:adaptive-kalman-filter}, which performs noise filtering without the explicit use of inputs $d_t$, $d_t^{(1)}$, and $d_t^{(2)}$.

\begin{algorithm}[H]
\caption{\hltext{IR with adaptive EnKF}}\label{alg:adaptive-kalman-filter}
\begin{algorithmic}
\STATE 
\STATE \textbf{Input: }$r_1^{(2)}, \dots, r_T^{(2)}$, $W_{\mcR}^{(1)}$, $\hat{f}_{\mcR}^{(1)}$, $\hat{Q}$, \hltext{$M$}, $\alpha_R$
\STATE \textbf{Output: }$r_1^{(2\to 1)}, \dots, r_T^{(2\to 1)}$
\STATE \hltext{$r_{0/0}^{\circledtiny{1}},\dots,r_{0/0}^{\circledtiny{$M$}} \overset{i.i.d.}{\sim} \mathcal{N}(0,I)$}, $R_0 \leftarrow I$
\FOR{$t = 1, \dots, T$}
\STATE \hltext{$r_{t/t-1}^{\circledtiny{$i$}} \overset{i.i.d.}{\sim} \mathcal{N}(\hat{f}_{\mcR}^{(1)}(r_{t-1/t-1}^{\circledtiny{$i$}}), Q)$ for each $i=1,\dots,M.$}
\STATE \hltext{$v_{t}^{\circledtiny{$i$}} \overset{i.i.d.}{\sim} \mathcal{N}(0, R_{t-1})$ for each $i=1,\dots,M.$}
\STATE \hltext{$U_{t/t-1}\leftarrow$ sample covariance matrix between $\{ r_{t/t-1}^{\circledtiny{$i$}} \}_{i=1}^M$ and $\{ r_{t/t-1}^{\circledtiny{$i$}}+v_{t}^{\circledtiny{$i$}} \}_{i=1}^M$.
}
\STATE \hltext{$V_{t/t-1}\leftarrow$ sample auto-covariance matrix of $\{ r_{t/t-1}^{\circledtiny{$i$}}+v_{t}^{\circledtiny{$i$}} \}_{i=1}^M$.}
\STATE \hltext{$K_t \leftarrow U_{t/t-1}V_{t/t-1}^{-1}$}
\STATE \hltext{$r_{t/t}^{\circledtiny{$i$}} = r_{t/t-1}^{\circledtiny{$i$}} + K_t(r_t^{(2)} - r_{t/t-1}^{\circledtiny{$i$}} - v_t^{\circledtiny{$i$}})$ for each $i=1,\dots,M.$}
\STATE \hltext{$r_t^{(2\to 1)}\leftarrow$ sample mean of $\{ r_{t/t}^{\circledtiny{$i$}} \}_{i=1}^M$.}
\STATE \hltext{$P_{t/t}\leftarrow$ sample auto-covariance matrix of $\{ r_{t/t}^{\circledtiny{$i$}} \}_{i=1}^M$.}
\STATE $v_t \leftarrow r_t^{(2)} - r_t^{(2\to 1)}$
\STATE $R_t \leftarrow (1-\alpha_R)\cdot R_{t-1} + \alpha_R\cdot(v_tv_t^\top + P_{t/t})$
\ENDFOR
\end{algorithmic}
\end{algorithm}

\subsubsection{Numerical Experiments}
We conducted numerical experiments to demonstrate IR with noise filtering using prior knowledge.
We considered the original input $d_t$, which is defined as follows:
\begin{alignat*}{1}
    d_t = \cos(2\pi t/100)\;(1\leq t \leq 5000).
\end{alignat*}
We generated inputs $d_t^{(1)} = d_t + \epsilon_t^{(1)}$ and $d_t^{(2)} = d_t + \epsilon_t^{(2)}$, where $\epsilon_t^{(1)}$ and $\epsilon_t^{(2)}$ are zero-mean white Gaussian noise with variance $\text{tr}(\Sigma^{(1)}) = 0.01$ and $\text{tr}(\Sigma^{(2)})$ is varied over 10 logarithmically spaced values between $10^{-2}$ and $10^1$.
We used a \hltext{30}-dimensional ESN with the element-wise $\tanh$ as the activation function $\sigma.$
We set the elements of the input layer parameter \hltext{$A\in\R^{30\times 1}$} using i.i.d samples obtained from $\mathcal{N}(0, 0.02)$.
Let \hltext{$B_0\in\R^{30\times 30}$} be a matrix with elements being i.i.d samples from $\mathcal{N}(0, 1)$.
We set \hltext{$B = 0.9\cdot\rho(B_0)\cdot B_0$}, where $\rho(B_0)$ is the spectral radius of $B_0$. 
For each ESN implementation and noisy inputs \hltext{$\mcD_{1,5000}^{(1)}$} and \hltext{$\mcD_{1,3000}^{(2)}$}, we calculated reservoir states \hltext{$\mcR_{1,5001}^{(1)}$} and \hltext{$\mcR_{1,3001}^{(2)}$} with the zero vector as the initial reservoir state.
We applied Algorithm \ref{alg:prior-knowledge} to obtain $W_{R}^{(1)}, \hat{f}_{\mathcal{R}}^{(1)}$ and $\hat{Q}$ from \hltext{$\mcR_{1,5001}^{(1)}$}.
Then, we applied the adaptive \hltext{EnKF} Algorithm \ref{alg:adaptive-kalman-filter} with \hltext{$\alpha_R = 0.01$ and $M = 300$} to obtain filtered reservoir states $r_{t}^{(2\to 1)}$ from the sequence of contaminated reservoir states $r_t^{(2)}$.
We conducted simulations for \hltext{10} different implementations of the ESN parameters $A$ and $B$ and noisy inputs \hltext{$\mcD_{1,5000}^{(1)}$} and \hltext{$\mcD_{1,3000}^{(2)}$}.
We calculated the relative root mean squared error (RRMSE) to evaluate the IR performance.

\hltext{
We additionally evaluated the proposed method under non-Gaussian observation noise, including heavy-tailed distributions such as the Student's $t$ and Cauchy distributions. Results demonstrating consistent performance are reported in the Supplementary Material.
}

Figure \ref{fig:num-filter1} shows a typical numerical example of the ESN inputs and outputs using IR with noise filtering.
We used low-intensity (Figure \ref{fig:num-filter1}(\textbf{a})) and high-intensity (Figure \ref{fig:num-filter1}(\textbf{b})) noise inputs for training and testing, respectively.
In this trial, the RRMSE of IR was reduced from \hltext{$1.03$ to $0.50$} by applying noise filtering (Figure \ref{fig:num-filter1}(\textbf{c}) and (\textbf{d})).

\begin{figure}[H]
\begin{center}
\includegraphics[width=110mm]{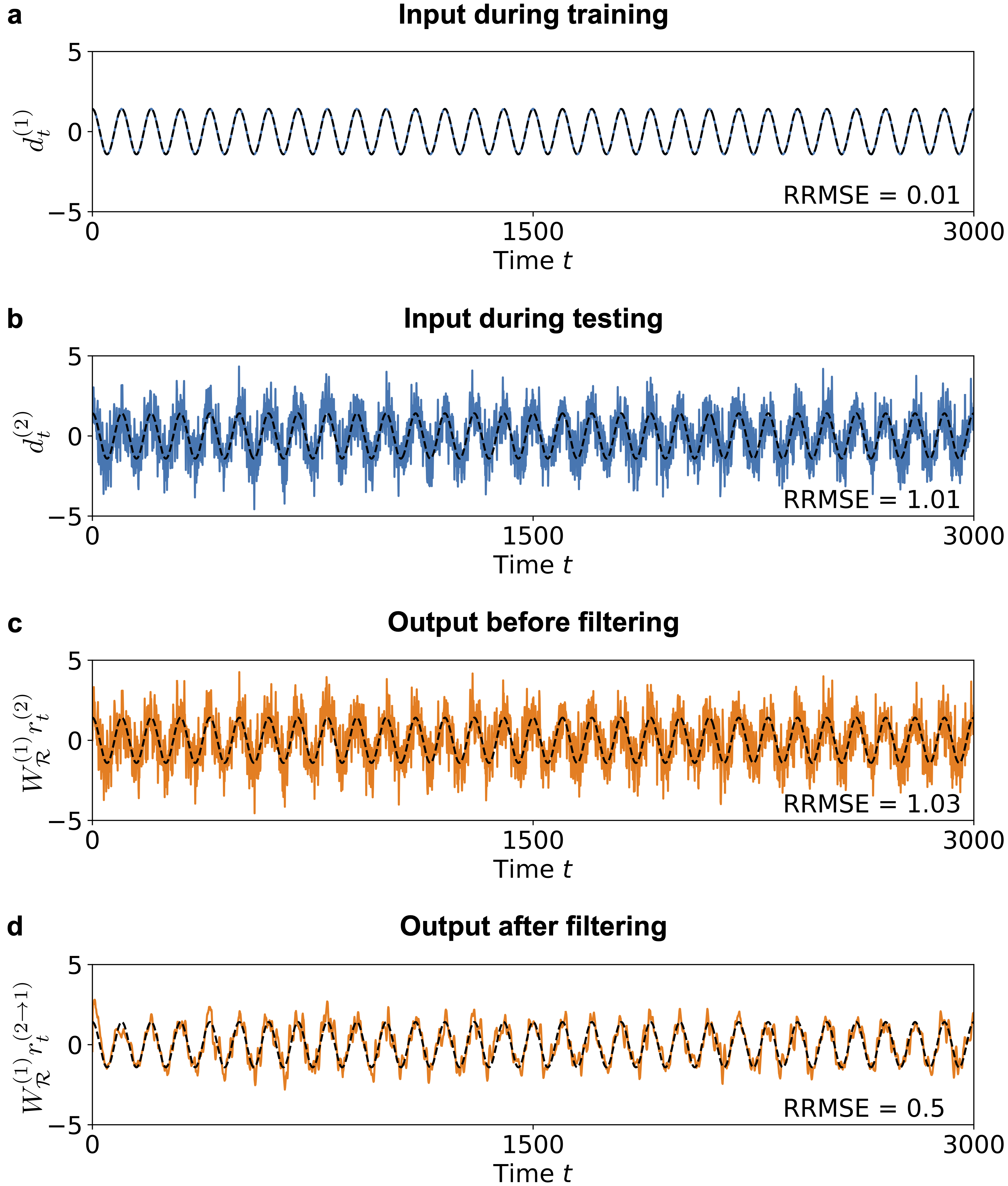}
\caption{
Typical example of input reconstruction with noise filtering.
\textbf{(a)} ESN input $d_t^{(1)}$ with low-intensity noise $(\text{tr}(\Sigma^{(1)}) = 0.01)$ used in prior training.
\textbf{(b)} ESN input $d_t^{(2)}$ with high-intensity noise $(\text{tr}(\Sigma^{(2)}) = 1)$ used in testing.
\textbf{(c)} ESN output $W_{\mcR}^{(1)}r_t^{(2)}$ before filtering.
\textbf{(d)} ESN output $W_{\mcR}^{(1)}r_t^{(2\to 1)}$ after filtering.
The original input $d_t$ is shown as a reference (black dashed line) in all panels.
}
\label{fig:num-filter1}
\end{center}
\end{figure}
\begin{figure}[H]
\begin{center}
\includegraphics[width=120mm]{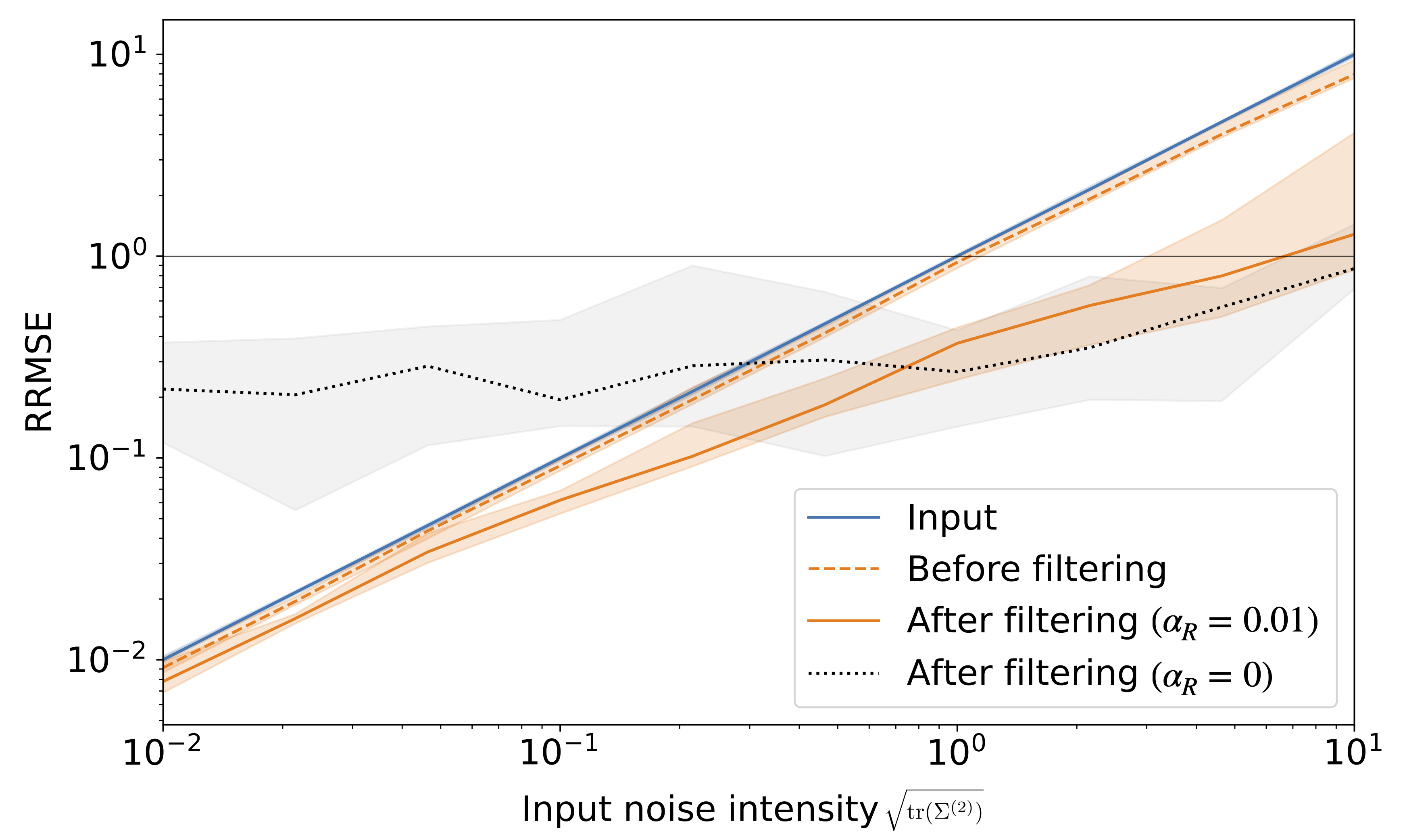}
\caption{Input reconstruction performance for different input noise intensities.
The means and standard deviations among 10 trials are depicted by the lines and shaded regions, respectively.
}
\label{fig:num-filter2}
\end{center}
\end{figure}

\newpage
Figure \ref{fig:num-filter2} shows the average behavior of the IR RRMSE across \hltext{10} trials for each noise intensity of inputs $d_t^{(2)}$ used during testing.
We observed that applying noise filtering decreases IR RRMSE on average when the input noise intensity during testing is higher than that during training (Figure \ref{fig:num-filter2}, solid orange and dashed orange lines).
We also found that adaptively updating the covariance estimate $\hat{R}$ (with $\alpha_R = \hltext{0.01}$) significantly lowers the RRMSE of input reconstruction when the test input is contaminated with low-intensity noise ($\sqrt{\text{tr}(\Sigma^{(2)})} < 1$).
Noise filtering without adaptation of covariance matrix $\hat{R}_t$ also works when the input noise intensity during testing $\sqrt{\text{tr}\Sigma^{(2)}}< 1$ is $10^{0}$ to $10^{1}$ due to the suitability of the initial covariance matrix estimate $\hat{R}_0$ to the noise level in the test data.

\section{Discussion}
\label{sec:dis}
\subsection{Unsupervised Input Reconstruction}  

In Section \ref{sec:uns}, we demonstrated that IR in ESN can be formalized as UL, which does not require data from the original inputs.  

\subsubsection{ESN information processing ability}

This UL-based formalization enhances our understanding of the ESN information processing ability.  
The ESN readout layer is trained to match the desired ESN output (see Equation \ref{eq:st-readout}).  
As the desired output is generally a nonlinear and time-delayed transformation of the input, the input itself can be regarded as the simplest choice for the desired output.  
In this sense, the IR performance could correspond to the upper limit of the ESN information processing ability.  
Our finding that IR in ESNs is UL indicates that the upper limit of the ESN information processing ability is independent of the type of input.  
This perspective allows us to independently investigate the following questions:
\begin{itemize}  
    \item What kind of ESN properties, regarding the activation function $\sigma$ and parameters $A$ and $B$, are suitable for IR?  
    \item Which properties of the ESN input degrade its raw information processing ability?  
\end{itemize}  
Thus, the UL-based IR formalization provides guidelines for understanding both the universal and task-dependent ESN properties.  

\subsubsection{ESN behavior toward the input}
We developed an algorithm to compute the readout layer parameter \( W_{\mcR} \in \mathbb{R}^{\nin \times \nr} \), which maps the reservoir state \( r_t \) to the current input \( d_t \).  
Using this, we can numerically construct a parametric family of candidate inverse maps of \( W_{\mcR} \) as follows:
\begin{alignat}{1}  
    \label{eq:linear-esm}  
    V_{\mcR, \Xi} := W_{\mcR}^+ + (I - W_{\mcR}^+ W_{\mcR})\Xi,  
\end{alignat}  
where \( \Xi \in \mathbb{R}^{\nr \times \nr} \) is an arbitrary matrix.
Suppose that \( \operatorname{rank}(W_{\mcR}) = \nin \), i.e., \( W_{\mcR} \) is full row rank.  
Then the Moore–Penrose inverse satisfies \( W_{\mcR} W_{\mcR}^+ = I_{\nin} \), and since \( W_{\mcR}(I - W_{\mcR}^+ W_{\mcR}) = 0 \), we obtain \( W_{\mcR} V_{\mcR, \Xi} = I_{\nin} \).  
That is, \( V_{\mcR, \Xi} \) is a right inverse of \( W_{\mcR} \).
Furthermore, since the readout layer \( W_{\mcR} \) is trained to minimize the reconstruction error of the input, we have \( d_t \approx W_{\mcR} r_t \) by construction.  
Under this assumption and the full row rank condition, it is plausible that \( r_t \approx V_{\mcR, \Xi} d_t \) for some choice of \( \Xi \).  
This suggests that \( V_{\mcR, \Xi} \) may serve as a linear surrogate for the inverse mapping from input to reservoir state.
From this perspective, \( V_{\mcR, \Xi} \) can be interpreted as a linear approximation of the echo state map \citep{hart-embedding-2020}, generalized synchronization \citep{kocarev1996generalized, lu-invertible-2020}, and embedding \citep{sauer_embedology_1991, grigoryeva_chaos_2021}, all of which relate the input signal \( d_t \) to the internal reservoir state \( r_t \).  
This connection provides a principled framework for analyzing how ESNs encode input signals into internal states through approximate inverse mappings.
In future work, the following issues need to be addressed:  
\begin{itemize} 
    \item How can we calculate matrix $\Xi$ in Equation \ref{eq:linear-esm} to satisfy $V_{\mcR, \Xi} d_t \approx r_t$?  
    \item Is the calculation of $\Xi$ possible using UL?  
\end{itemize}

\subsection{Unsupervised Dynamical system Replication}  

In Section \ref{subsec:dyn-rep}, we demonstrated that the replication of a dynamical system generating the ESN input can be achieved using UL.  

\subsubsection{Standard approach to dynamical system replication}
The UL-based formalization encourages us to reconsider the standard approach to dynamical system replication.  
We observed that the replicated dynamical systems using SL and UL differ because of numerical errors.  
These errors are practically unavoidable.  
For example, numerical errors in the computation of the inverse tanh function are unavoidable because they involve Napier's constant, which is approximated as a rational number in computers.  
Similarly, numerical errors in the computation of the Moore-Penrose inverse of matrices are unavoidable because of the singular value decomposition, where singular values smaller than the machine epsilon are typically truncated by default.  
Thus, we propose treating the replicated dynamical systems obtained from SL and UL as distinct, even though they are theoretically equivalent, based on assumptions made for ESNs, as explained in Section \ref{sec:uns}.  
Moreover, we propose considering the replicated dynamical systems obtained from UL as a standard because this approach does not require data for the original ESN input.  
In contrast, the replicated dynamical systems obtained from SL are treated as special cases that can be implemented only when the original ESN input is exceptionally available.  
It is worth investigating how the assumption of the original input availability affects characteristics such as dynamical system replication robustness.  

\subsubsection{Relations to theoretical neuroscience}
A formalization independent of the original ESN input enables the investigation of the computational principles of self-organizing systems, such as the brain.  
Our perception, sustained by the brain, allows us to understand the external environment through sensory information.  
All computations required for perception occur within the brain and do not explicitly depend on the true state of the external environment.  
In this context, our UL-based formalization contributes to further exploration of the computational principles of the brain.  
In particular, if the activation function $\sigma$ is element-wise and injective, such as the tanh or the identity function, the loss function defined in Equation \ref{eq:loss} is equivalent to the following loss function of $\hat{B}$:  
\begin{alignat}{1}  
    \label{eq:loss-nonlinear}  
    \sum_{t=1}^{T-1}\| \underbrace{r_{t+1}}_{\text{stimulus-evoked activity}} - \underbrace{\sigma(\hat{B}r_t)}_{\text{internally-predicted activity}}\|^2.  
\end{alignat}  
This reformulation interprets the loss function $L_{\sigma, T}(\hat{B})$ as the discrepancy between the stimulus-evoked activity $r_{t+1}$ and the internally-predicted activity $\sigma(\hat{B}r_t)$.  
This perspective parallels prior studies \citep{rao_predictive_1999, knill_bayesian_2004, friston_free-energy_2010, asabuki_embedding_2024}, which propose that minimizing the discrepancy between these activities reflects a core computational principle underlying perception.  
We highlight a shared mathematical foundation between the prior studies addressing brain models for perception and the ESN's computational framework.  
Notably, we deductively derived this loss function from the mathematical observation on IR, hence avoiding heuristic implementation of the loss function \ref{eq:loss-nonlinear}. 
Provided that ESN is one of the simplest models capturing the essence of perception based on the loss function \ref{eq:loss-nonlinear}, we believe the ESN potential as a theoretical model of the brain.
To the best of our knowledge, it remains unclear whether the introduction of additional complexities in the model guarantee qualitative benefits in the tasks treated in this study.
We believe that our formulation using ESN provides a foundational basis for future investigations into the detailed properties of both the brain and ESNs.

\subsection{Unsupervised Noise Filtering}  

In Section \ref{subsec:filtering}, we demonstrated that a replicated dynamical system can be used for noise filtering, and the algorithm to achieve this does not require explicit use of the ESN input.  

\subsubsection{Effective design of ESN with error feedback}
We proposed a method that fully applies the Kalman filter algorithm to ESNs. 
This approach induces the design of an ESN with effectiveness being supported by the Kalman filter theory.  
We consider a linear ESN, where the activation function $\sigma$ is the identity function.  
Under this assumption, we obtain the asymptotic form of the dynamics of the filtered reservoir state $r_{t}^{(2\to 1)}$ 
\begin{alignat}{1}  
\label{eq:filter-dyn}  
    r_{t+1}^{(2\to 1)} &= \hat{f}^{(1)}_{\mcR}( r_{t}^{(2\to 1)} ) + K(r_{t+1}^{(2)} -\hat{f}^{(1)}_{\mcR}( r_{t}^{(2\to 1)} )),  
\end{alignat}  
where $K = P(P+R)^{-1} \in \R^{\nr \times \nr}$ is the stationary Kalman gain, defined using the solution $P$ of the algebraic Riccati equation $P = \hat{B}(P  - P(P+R)^{-1}P)\hat{B}^\top + Q$.
Equation \ref{eq:filter-dyn} is reformulated as follows:  
\begin{alignat}{1}  
\label{eq:gen-error-feedback}
\nonumber
r_{t+1}^{(2\to 1)}  
    =& \underbrace{g(d_t^{(2)}, r_{t}^{(2\to 1)})}_{\text{main dynamics}}\\\nonumber 
    &+ \underbrace{(K-I)\left[  
g(d_t^{(2)}, r_{t}^{(2\to 1)}) - g(W_{\mcR}r_{t}^{(2\to 1)}, r_{t}^{(2\to 1)})  
\right]}_{\text{error feedback}} \\
    &+ \underbrace{\left[  
g(d_t^{(2)}, r_{t}^{(2)}) -  
g(d_t^{(2)}, r_{t}^{(2\to 1)})  
\right]}_{\text{residual}}.  
\end{alignat}  
We note that Equation \ref{eq:gen-error-feedback} can be viewed as an ESN with IR error feedback \citep{katori_network_2018} defined as follows:  
\begin{alignat}{1}  
    r_{t+1} := \underbrace{g(d_t^{(2)},  r_{t} )}_{\text{main dynamics}} + \underbrace{W^{\text{fb}}(d_t^{(2)} - W_{\mcD}r_{t})}_{\text{error feedback}},  
\end{alignat}  
where $W^{\text{fb}}\in\R^{\nr\times\nr}$ is the ESN feedback layer.  
Thus, the proposed noise filtering algorithm motivates the design of ESNs incorporating IR error feedback. 
In particular, parameter $W^{\text{fb}}$ in the ESN feedback layer recommended to be set as $(K-I)A$, provided that the linearity assumption of the activation function $\sigma$ is acceptable.  
This parameter setting, which uses the Kalman gain $K$, implements an intuitively natural strategy: it utilizes information obtained from current input $d_t^{(2)}$ only if its noise intensity is smaller than the learning error in the readout layer $W_{\mcR}$.  
Therefore, the construction of the filtering algorithm without the explicit use of the ESN input theoretically supports the benefits of error feedback in ESNs, as reported, in \citep{yonemura2021network, sato_silico_2024}, for example. 

\hltext{
\subsubsection{Extensions toward Non-stationary Noise}

In filtering problems, the system noise covariance matrix $Q$ and the observation noise covariance matrix $R$ are typically incorporated into the algorithm to reflect the probabilistic characteristics of the system state and observations.

In the state-space model introduced in this study (Equation~\ref{eq:state-space-model}), system noise is formulated as the modeling error of a dynamical system that has been pretrained in an unsupervised manner using an ESN. Accordingly, the matrix $Q$ is fixed once pretraining is completed and does not vary with the noise characteristics of the input sequence. In contrast, observation noise is defined as the discrepancy between the actual ESN state and its estimate at each time step, and thus the matrix $R$ is influenced by the noise properties of the ESN input and cannot be determined a priori.

This study focuses on a stationary setting in which the true observation noise covariance is unknown but constant. Under this assumption, we numerically verified the effectiveness of adaptive filtering for online estimation of $R$ (Figure~\ref{fig:num-filter2}, solid orange and dashed orange lines). More generally, input noise may be non-stationary. The adaptive filter~\citep{mehra_approaches_1972} employed in this work is one standard approach to address such cases. The observed improvement from adaptation (Figure~\ref{fig:num-filter2}, solid orange and dotted black lines) suggests that the proposed algorithm is capable of estimating unknown and potentially time-varying covariance matrices, and therefore remains effective as long as the temporal variation of noise is sufficiently slow.

Alternative approaches to non-stationary noise filtering include the Switching Kalman Filter~\citep{murphy1998switching}, which simultaneously runs multiple filters based on different state-space models, and Augmented State Kalman Filtering~\citep{deaves1999covariance}, which assumes prior knowledge of the covariance dynamics and incorporates them as part of the system state. Extending our framework to incorporate such methods may be a fruitful direction for future development, depending on the characteristics of the target application.
}

\subsection{Limitations}

\subsubsection{Theoretical requirements for enabling UL formulation of IR in ESNs}

This study has clarified the theoretical assumptions on ESN parameters required to enable UL formulation of IR within ESNs. 
These assumptions fall into two main categories: (1) regularity conditions on the ESN and (2) full accessibility of the internal structure of the ESN.

The first requirement concerns the full column rank dondition of the input layer matrix \( A \) and the invertibility of activation function \( \sigma \), which enables the reconstruction of the input time series \( \mcD_{1,T} \) from the internal ESN state \( \mcR_{1,T+1} \).  
While these conditions may seem restrictive in theory, they are often satisfied in practice when using randomly generated input matrices and standard activation functions such as the identity or hyperbolic tangent function, both of which are commonly employed in ESN implementations. 
\hltext{
We numerically confirmed that the non-invertibility of activation functions degrades the performance of IR, with ReLU-based ESNs achieving an RRMSE of approximately $10^{-2}$ (Figure \ref{fig:S2-RRMSE}).
Although this is worse than the near-perfect performance under invertible functions, it remains significantly better than chance level (RRMSE = 1).
Accordingly, we suggest that nonlinear invertible activation functions, such as the hyperbolic tangent, are more tractable than non-invertible ones under the proposed theory of IR in ESN.
} 

The second requirement is full knowledge of the internal components of the ESN, specifically \( A \), \( B \), and \( \sigma \).  
This assumption allows us to reformulate IR as a UL problem by substituting the need for target outputs with knowledge of the model parameters.
\hltext{
The practicality of this assumption depends on the specific application scenario.
As highlighted in this study, digital reservoirs such as ESNs typically provide full access to internal structures, including connectivity matrices and activation functions.
Our findings support the utility of leveraging such prior information, as they demonstrate that IR, which is typically formulated as an SL task, can instead be reformulated as a UL problem when the internal components are known.
In contrast, physical reservoir systems \citep{tanaka_recent_2019}, including analog circuits \citep{appeltant_information_2011}, soft robotic systems \citep{nakajima_information_2015}, and chemical reaction systems \citep{baltussen_chemical_2024}, often operate under practical constraints that limit precise design or access to internal dynamics.
Historically, the reservoir computing paradigm has implicitly treated the reservoir as a black-box component due to its fixed and untrained nature \citep{jaeger2001echo, maass-real-time-2002}. 
As a result, explicit modeling of the reservoir’s internal dynamics has received little attention.
However, this perspective tends to conflate two distinct assumptions: that the reservoir is untrained, and that its internal structure is inaccessible. While the former remains a central tenet of reservoir computing, the latter is not inherently necessary.
In this context, the present work provides a new perspective, pointing toward research directions that take advantage of the internal behavior of physical reservoirs.
While specifying exact connectivity matrices or activation functions in such systems may be challenging, reproducibility of internal dynamics across fabricated instances may still be attainable.
For instance, analog circuits are typically engineered to behave deterministically, apart from variability due to manufacturing tolerances or thermal noise.
Soft robots exhibit structural consistency in their dynamics once fabricated, and chemical reaction systems are constrained by physical principles such as conservation of mass, which impose qualitative regularities.
The central question is whether one can construct a reliable dynamical model that captures these deterministic features.
Given recent advances in the simulation of physical systems, we believe that our UL-based formulation has the potential to be extended to the domain of physical reservoir computing.

An even more challenging setting arises when the identity and structure of the reservoir system are entirely unknown.
In such cases, our formulation introduces a novel system identification problem: inferring the reservoir model solely from observed reservoir state trajectories.
Although we do not propose a specific solution strategy in this study, several theoretical challenges are evident.
Assuming that the target reservoir can be approximated by a suitable ESN, the problem becomes one of estimating the activation function and weight matrices from the observed reservoir states.
This raises fundamental issues such as identifiability, for example, scale or permutation indeterminacy in the input weight matrix when the input time series is unobserved.
It also introduces an infinite-dimensional estimation problem due to the unknown activation function, placing the task within the domain of semi-parametric inference \citep{li_regression_1989, bickel1993efficient}.
Determining which classes of ESNs are identifiable from reservoir state sequences alone, and whether such classes retain expressive capabilities such as universal approximation, represent promising avenues for future research suggested by our formulation.
}

\subsubsection{Performance evaluations}

This study does not include a performance comparison between our proposed method and existing algorithms for IR, such as those developed in the context of blind source separation \citep[e.g.,][]{bell1995information, amari_adaptive_1998, isomura_local_2016, asabuki_somatodendritic_2020, bahroun_normative_2021}, dynamical system replication using machine learning models other than RC \citep[e.g.,][]{chen2018neural, dueben2018challenges, scher2019generalization}, or model-free noise filtering \citep[e.g.,][]{vincent2008extracting, hamilton2016ensemble}.  
This omission is intentional: our primary aim is not to develop a new state-of-the-art algorithm, but to establish a theoretical framework for UL-based IR and its connections to related tasks.
Our contribution lies in identifying the conditions under which IR can be achieved without supervision in ESNs, and in demonstrating that related tasks such as dynamical system replication and noise filtering can be reformulated within this unified framework.  
These reformulations enable the transfer of existing theoretical guarantees into the UL setting.

Nonetheless, we acknowledge that empirical evaluation on real-world tasks is essential for assessing practical utility.  
Future work should explore the performance and limitations of UL-based IR, particularly under noisy or data-constrained conditions.

\section{Conclusion}
\label{sec:con}

In this study, we demonstrated that input reconstruction in echo state networks can be formulated as an unsupervised learning task.  
Our findings highlight that knowledge of the specific values of fixed parameters in echo state networks is beneficial for both theoretical formalization and practical implementation.  
Accordingly, we propose a new principle for reservoir computing: not only should certain parameters be fixed, but their specific values should also be actively exploited.  
We also presented methods for dynamical system replication and noise filtering based on input reconstruction through unsupervised learning.  
These results support a broader perspective on echo state networks, reinforcing their status as canonically multifunctional systems and underscoring their potential as theoretical models of complex systems such as the brain.

\section*{Acknowledgments}
This work was partially supported by JSPS KAKENHI Grant Numbers JP20H00596, JP21K12105, JP22K18419, JP24K15161, JP24H02330, JST CREST Grant No. JPMJCR19K2, JST Moonshot RD Grant No. JPMJMS2021, Cross-ministerial Strategic Innovation Promotion Program (SIP) on “Integrated Health Care System” Grant No. JPJ012425.

{
\bibliographystyle{apalike}
\bibliography{
ref,
RCforDynamicsLearning,
KalmanFilterTheory,
AdaptiveKalmanFilter,
PredRC,
ESNmap-embedding-GS,
PredictiveTheory,
RC-review,
BSS
}
}

\hltext{
\clearpage
\renewcommand{\thefigure}{S\arabic{figure}}
\setcounter{figure}{0}
\section*{Supplementary Material for \\
Unsupervised Learning in Echo State Networks
for Input Reconstruction}


\subsection*{Input Reconstruction under Violations of Regularity Conditions}

Our theory of UL in ESN for IR relies on three independent regularity conditions:  
(i) invertibility of the activation function $\sigma$,  
(ii) full column rank of the input weight matrix $A$, and  
(iii) full row rank of the reservoir state matrix $\mcR_{1,T}$.

As mentioned in Section~\ref{subsec:IR}, condition (ii) is practically satisfied, as a randomly generated matrix is full rank with high probability.  
In practice, ill-conditioned scenarios related to conditions (i) and (iii) may realistically occur depending on ESN settings.  
Here, we provide empirical insights into the feasibility of such cases for UL in ESN for IR.

\subsubsection*{Input Reconstruction with Non-Invertible Activation Functions}

We discuss the feasibility of using non-invertible activation functions for UL in ESN for IR.  
Specifically, we consider an ESN with the ReLU activation function:
\begin{alignat}{1}
    r_{t+1} = \text{ReLU}(Ad_t + Br_t),
\end{alignat}
where $\text{ReLU}(x) = \max(0,x)$~\citep{glorot2011deep}.  
ReLU is non-invertible because it maps all non-positive values to zero.  
The pre-activation value $m_t = Ad_t + Br_t$ can take non-positive values due to possibly non-positive entries in $A$, $B$, or $d_t$, even though $r_t$ is non-negative.  
Consequently, the ReLU function transforms $m_t$ into $r_{t+1}$ while discarding information about $m_t$, and therefore $d_t$ (Figure~\ref{fig:S1-hist}).  
In short, a unique inverse of ReLU does not exist.  

To apply our UL method for IR in such settings, a surrogate inverse for ReLU must be introduced.  
Here, we consider two implementations of surrogate inverses: $\text{ReLU}_{\alpha}^{-1}$ and $\text{ReLU}_{\text{random}}^{-1}$, defined as follows:
\begin{alignat}{1}
    \text{ReLU}_{\alpha}^{-1}(x) &= \begin{cases}
        x & (x \geq 0)\\
        \alpha & (x < 0)
    \end{cases},\\
    \text{ReLU}_{\text{random}}^{-1}(x) &= \begin{cases}
        x & (x \geq 0)\\
        \xi \sim \text{dist}(-r_t) & (x < 0)
    \end{cases},
\end{alignat}
where $\text{dist}(-r_t)$ denotes the empirical distribution of the non-zero elements of $-r_t$.

Intuitively, $\text{ReLU}_{\alpha}^{-1}$ replaces the undefined negative domain with a fixed constant $\alpha < 0$ (Figure~\ref{fig:S1-hist}, bottom left), whereas $\text{ReLU}_{\text{random}}^{-1}$ samples a plausible negative value from the empirical distribution of the negative reservoir states (Figure~\ref{fig:S1-hist}, bottom right).

\begin{figure}[H]
\begin{center}
\includegraphics[width=120mm]{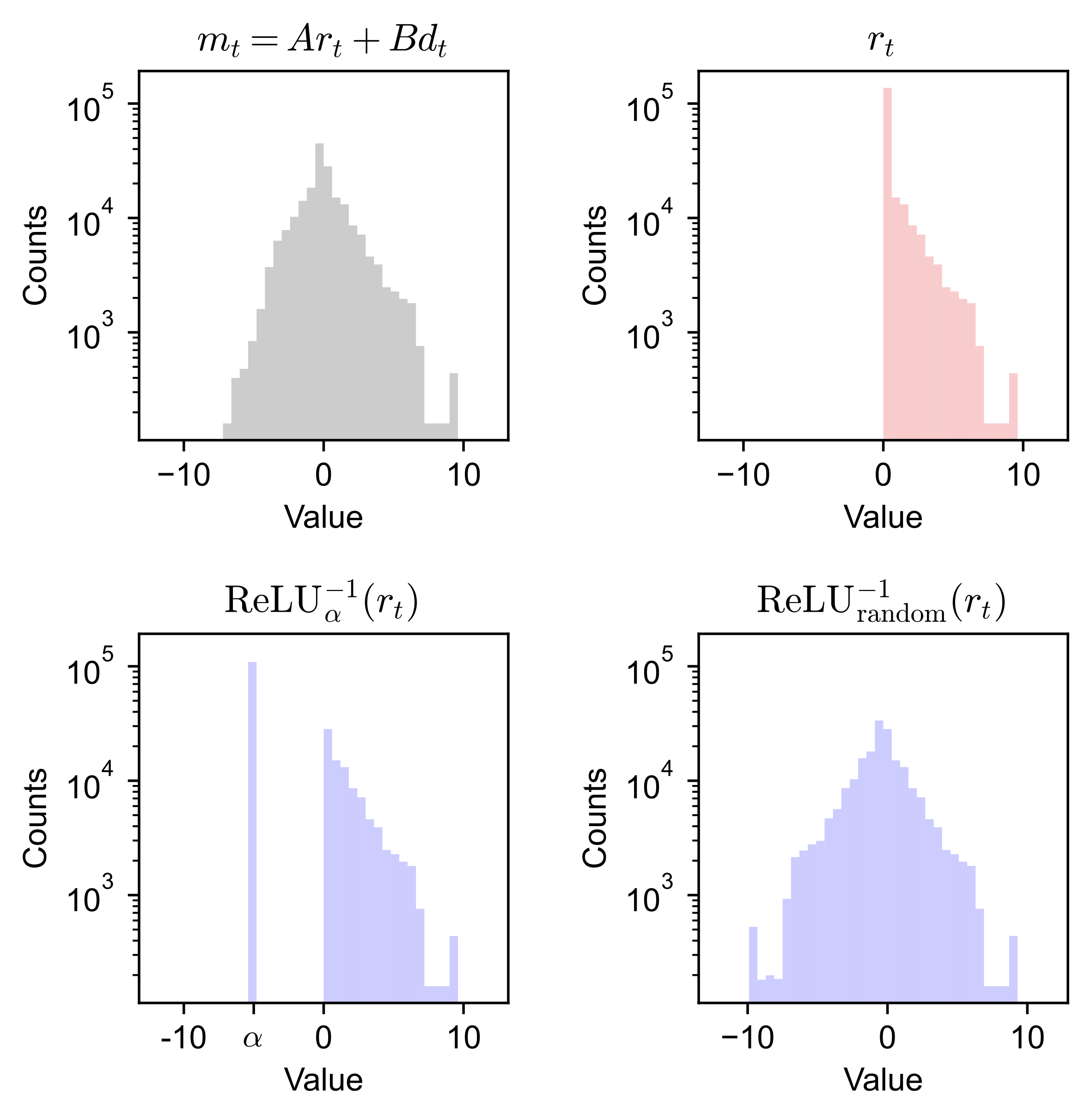}
\caption{
Histograms of the elements of the pre-activation state $m_t$, the ReLU-transformed state $r_t$, and their surrogate inverses.  
Top-left: $m_t = Ar_t + Bd_t$ (gray).  
Top-right: $r_t = \mathrm{ReLU}(m_t)$ (red), where negative values are mapped to zero.  
Bottom-left: $\mathrm{ReLU}_{\alpha}^{-1}(r_t)$ (blue), where zero-valued elements are replaced with a fixed constant $\alpha < 0$.  
Bottom-right: $\mathrm{ReLU}_{\mathrm{random}}^{-1}(r_t)$ (blue), where zero-valued elements are randomly sampled from the empirical distribution of non-zero elements of $-r_t$.
}
\label{fig:S1-hist}
\end{center}
\end{figure}

We evaluated the surrogate inverse of ReLU $\text{ReLU}_{\alpha}^{-1}$ and $\text{ReLU}_{\text{random}}^{-1}$ by RRMSE in IR.
In this experiment, the readout matrix $W_{\mathcal{R}}$ was trained in an unsupervised manner using Equation~\ref{eq:Wr}, and the ESN output $W_{\mathcal{R}} r_t$ was used as the prediction of the input $d_t$.
For ESN, we considered a 1-dimensional input $d_t = \cos(\pi t/50)$.
We used a 50-dimensional ESN with the element-wise ReLU as the activation function $\sigma$.
Except for the activation function and the initialization of input weight $A$, the setup was identical to the numerical experiments in Section\ref{subsec:IR}.
In particular, the reservoir matrix $B$ was constructed as $B = 0.9 \cdot \rho(B_0) \cdot B_0$, where $B_0 \in \mathbb{R}^{50 \times 50}$ is a random matrix with i.i.d. elements from $\mathcal{N}(0,1)$ and $\rho(B_0)$ denotes its spectral radius.
We examined the cases where the elements of the input weight matrix $A$ were drawn i.i.d. from zero-mean normal distributions with variances $0.01$, $1$, or $2$.
For each variance setting, we conducted numerical experiments across 10 independent realizations of $A$ and $B$.

Figure~\ref{fig:S1-RRMSE} shows the RRMSE of IR using the surrogate inverses of ReLU, $\text{ReLU}_{\alpha}^{-1}$ and $\text{ReLU}_{\text{random}}^{-1}$, under varying variances of the elements in the input matrix $A$.
The performance of both surrogate inverse approaches (RRMSE $\approx 10^{-2}$; Figure \ref{fig:S1-RRMSE}, solid and dashed blue lines) was worse than that of the SL method, which directly uses the input sequence $d_t$ (RRMSE $\approx 10^{-10}$; Figure \ref{fig:S1-RRMSE}, dotted blue line).
Nevertheless, it outperformed the chance level (RRMSE = 1; Figure \ref{fig:S1-RRMSE}, dotted black line), indicating that meaningful IR is still achievable even with a non-invertible activation function such as ReLU.

The optimal value of $\alpha$ for $\text{ReLU}_{\alpha}^{-1}$ varied depending on the scale of the input matrix (Figure~\ref{fig:S1-RRMSE}, solid blue lines).
This is because increasing the scale broadens the distribution of the pre-activation value $m_t$, shifting the mean and median of the values that are mapped to zero.
At the optimal $\alpha$, the IR performance was comparable to that of $\text{ReLU}_{\text{random}}^{-1}$ (Figure~\ref{fig:S1-RRMSE}, solid and dashed blue lines).
Thus, given the impracticality of determining the optimal $\alpha$ apriori, $\text{ReLU}_{\text{random}}^{-1}$ offers a more realistic and practical strategy for constructing a surrogate inverse.

It should be noted, however, that the validity of $\text{ReLU}_{\text{random}}^{-1}$ relies on the assumption that the distribution of $m_t$ is symmetric around zero.
This implies that some bias remains uncorrected, but such bias is an unavoidable consequence of the non-invertibility of ReLU.

\begin{figure}[H]
\begin{center}
\includegraphics[width=140mm]{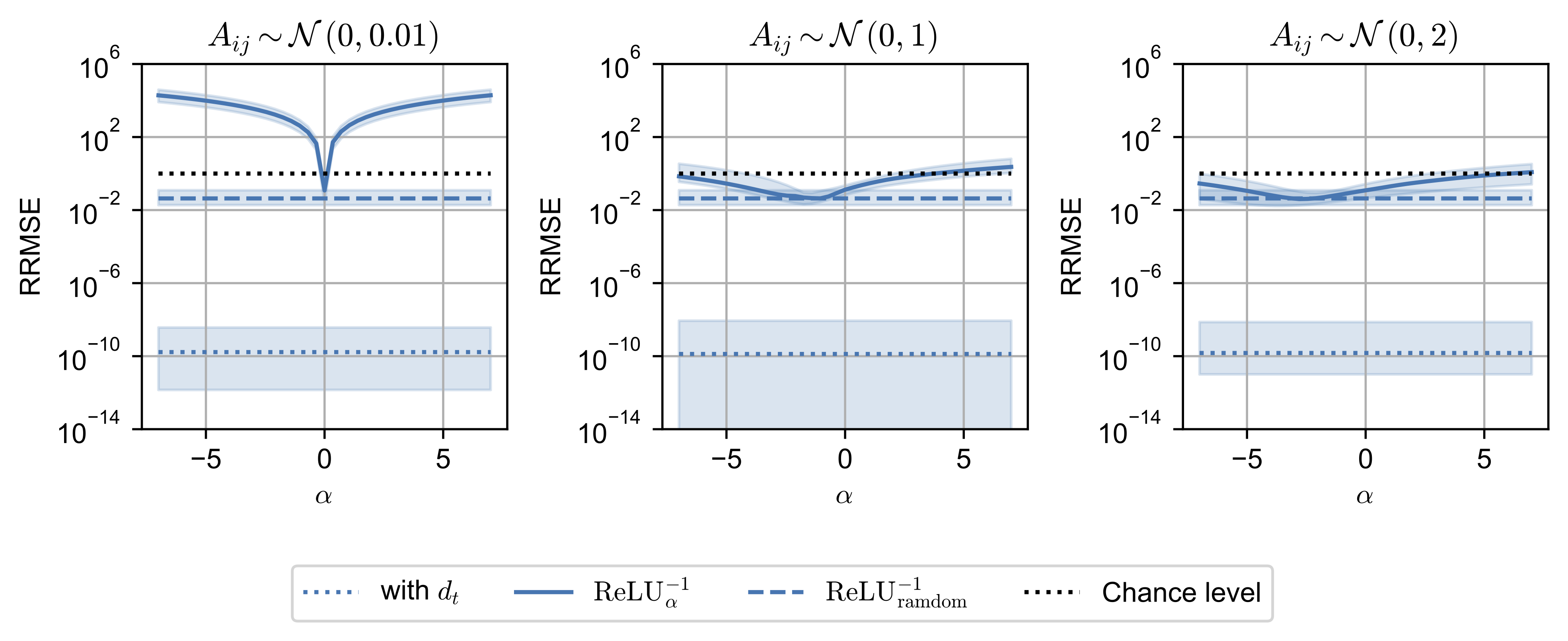}
\caption{
Input reconstruction performance under non-invertible activation functions.
Each panel shows the RRMSE of input reconstruction using surrogate inverses of the ReLU activation function: $\text{ReLU}_{\alpha}^{-1}$ (solid blue) and $\text{ReLU}_{\text{random}}^{-1}$ (dashed blue).
The input weight matrix $A$ is initialized with i.i.d. entries $A_{ij} \sim \mathcal{N}(0, \sigma^2)$, with three different variances:
Left: $\sigma^2 = 0.01$; Center: $\sigma^2 = 1$; Right: $\sigma^2 = 2$.
Dotted black lines indicate chance level (RRMSE = 1), and dotted blue lines show the performance of the supervised learning method using the original input $d_t$.
Shaded areas represent the range between the minimum and maximum RRMSE across 10 independent trials.
}
\label{fig:S1-RRMSE}
\end{center}
\end{figure}

\newpage
\subsubsection*{Input Reconstruction with Rank-Deficient Reservoir State Matrices}

Next, we discuss the feasibility of using rank-deficient reservoir state matrices in UL in ESN for IR.
Recall that the reservoir state matrix $\mcR_{1,T}$ is formed by concatenating $\nr$-dimensional column vectors $r_1, \dots, r_T$.
We consider the case $T \gg \nr$ and focus on the deficiency of full row rank in $\mcR_{1,T}$.

A row rank-deficient reservoir state matrix $\mcR_{1,T}$ was obtained using a periodic input and an ESN with the echo state property (ESP).  
To further control the degree of ill-conditioning, we added i.i.d.\ Gaussian noise to the input, defining it as  
\( d_t = \sin(\pi t / 50) + \epsilon_t \), where \( \epsilon_t \sim \mathcal{N}(0, \sigma^2) \).  
We varied $\sigma$ over a wide range using logarithmic scaling as numpy.logspace(-15, 0, 20).  
For each value of $\sigma$, we computed the rank of the corresponding reservoir state matrix $\mcR_{1,T}$ using numpy.linalg.matrix\_rank(),  
allowing us to empirically assess how the conditioning of $\mcR_{1,T}$ affects the performance of IR.

Specifically, we used a 1-dimensional input and a 100-dimensional ESN with the element-wise hyperbolic tangent function as the activation function $\sigma$.  
We initialized the elements of the input layer $A$ using i.i.d. samples from $\mathcal{N}(0, 0.02)$.  
Let $B_0 \in \R^{100 \times 100}$ be a matrix with elements being i.i.d. samples from $\mathcal{N}(0, 1)$.  
We set $B = 0.9 \cdot \rho(B_0) \cdot B_0$, where $\rho(B_0)$ denotes the spectral radius of $B_0$. 

Even in ESNs that satisfy ESP, the reservoir state \( r_t \) is influenced by initial transient dynamics, which can regularize the rank of the reservoir state matrix.
To mitigate this effect, we initialized the reservoir state with the zero vector and used the state obtained after 5000 time steps of evolution as the starting state for the experiment.

We then set the training data length to \( T = 5000 \), computed the readout weights for IR using the supervised method (Equation \ref{eq:input-rec-readout} and the unsupervised methods (Equation \ref{eq:Wr-1}, \ref{eq:Wr}), and evaluated the performance using RRMSE.

Figure~\ref{fig:S2-RRMSE} shows the relationship between the rank of the reservoir state matrix $\mcR_{1,T}$, the reconstruction error (RRMSE), and the noise intensity added to the input time series.  
First, we confirmed that adding noise to the input effectively eliminates the rank deficiency of the reservoir state matrix as intended (Figure \ref{fig:S2-RRMSE}, Top panel).  
Regularization through small amounts of noise provides a practical solution to improve the conditioning of $\mcR_{1,T}$, and a more detailed analysis of this approach is presented in \citep{wikner2024stabilizing}.

We highlight the following three key observations regarding
the behavior of RRMSE in IR (Figure \ref{fig:S2-RRMSE}, bottom panel):

First, the RRMSE achieves its minimum near the noise level at which the reservoir state matrix becomes full-rank.  
This observation aligns with a fundamental principle in reservoir computing: the diversity and complexity of reservoir states play a crucial role in effective information processing \citep{jaeger2001echo}.

Second, even when the noise level is low and the reservoir state matrix remains ill-conditioned, both the SL and the proposed UL methods still outperform the chance level (RRMSE = 1).  
This result indicates that the proposed approach possesses a degree of robustness in the face of ill-conditioning.

Third, and most notably, around the optimal noise level and under ill-conditioned settings, the proposed UL solution $W_{\mcR}$ (Equation~\ref{eq:Wr}) outperforms its intermediate form $W_{\mcR}^{(1)}$ (Equation~\ref{eq:Wr-1}), which is theoretically equivalent to the SL solution in exact arithmetic.
This advantage is likely due to numerical instability due to inversion order.  
Although it remains an open question whether this advantage is universal, the result suggests that the UL solution may offer a practical numerical benefit over the SL formulation in the IR setting.

\begin{figure}[H]
\begin{center}
\includegraphics[width=130mm]{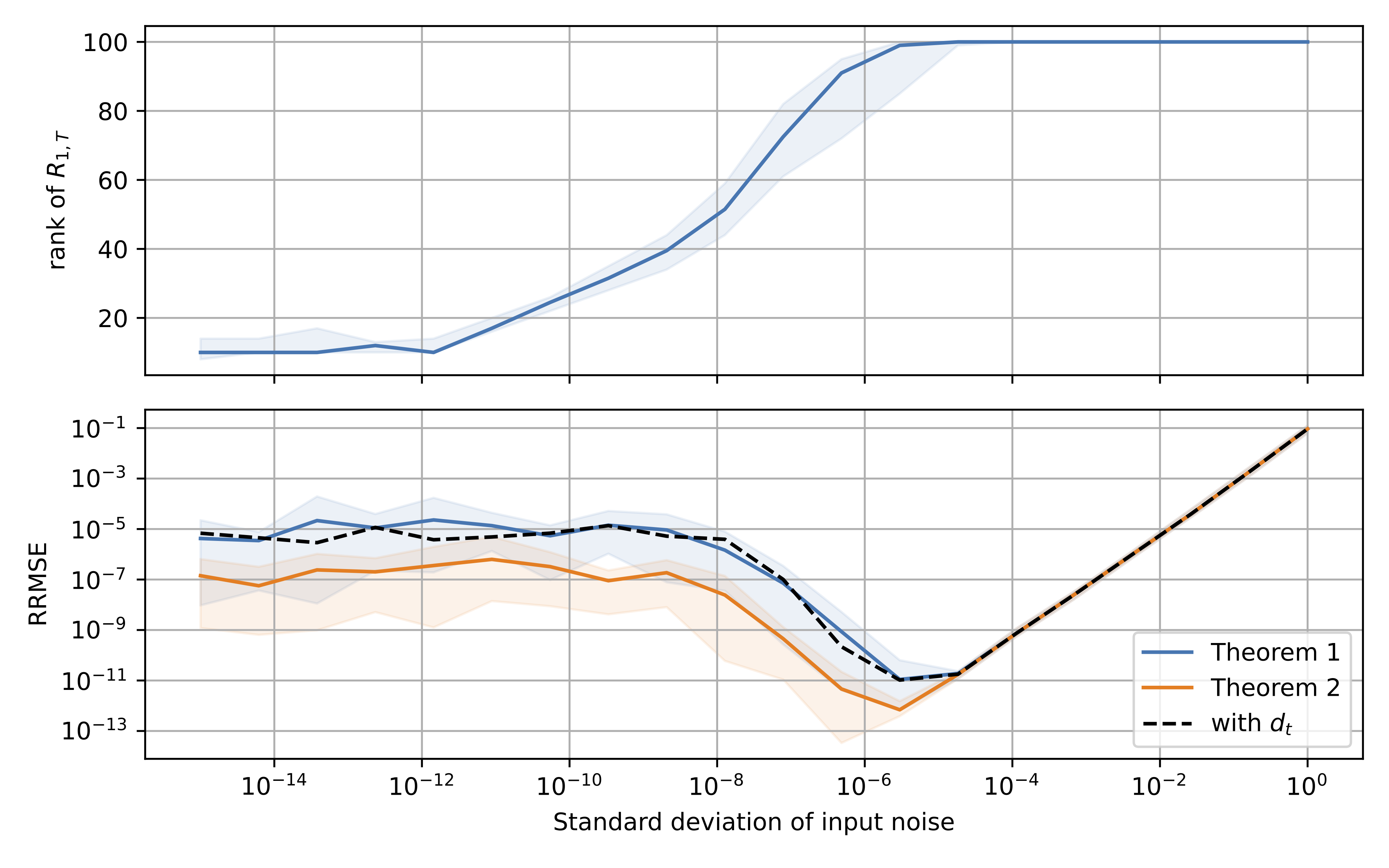}
\caption{
Effect of input noise on the rank of the reservoir state matrix and the accuracy of input reconstruction.
Both panels share the same horizontal axis, representing the standard deviation $\sigma$ of Gaussian noise added to the input sequence.
(Top) Rank of the reservoir state matrix $\mcR_{1,T}$, computed using numpy.linalg.matrix\_rank.
(Bottom) Reconstruction error (RRMSE) using the unsupervised solutions $W_{\mcR}^{(1)}$ (blue; Theorem \ref{key-thm}) and $W_{\mcR}$ (orange; Theorem \ref{key-thm-cor}), and the supervised solution (black dashed; Equation~\ref{eq:input-rec-readout}) using the true input $d_t$.
Shaded regions indicate the range between the minimum and maximum values across 10 trials.
}
\label{fig:S2-RRMSE}
\end{center}
\end{figure}

\subsection*{Robustness of Filtering-Based Input Reconstruction under Non-Gaussian Observation Noise}

In this study, we proposed an input reconstruction (IR) method that incorporates noise filtering to mitigate the impact of observation noise (Algorithms~\ref{alg:prior-knowledge} and~\ref{alg:adaptive-kalman-filter}).  
While the numerical experiments in the main text assumed Gaussian observation noise, it is practically important to assess the robustness of the filtering algorithm under non-Gaussian noise, which is common in real-world scenarios.

Although the filtering algorithm itself is based on the standard Ensemble Kalman Filter (EnKF), the state-space model used for filtering (Equation~\ref{eq:state-space-model}) is uniquely constructed via UL in ESN, as proposed in this study.  
Therefore, it is both theoretically and practically significant to evaluate whether this formulation remains effective under non-Gaussian noise conditions.

To assess robustness under such conditions, we repeated the input filtering experiment with two modifications: the ESN dimension was increased from 30 to 50, and the ensemble size $M$ in the EnKF from 300 to 500.  
Specifically, we replaced the white Gaussian noise added to the test-time ESN input $d_t^{(2)}$ with noise drawn from the Student’s $t$ distribution, and evaluated the RRMSE of IR.  
The Student’s $t$ distribution is parameterized by degrees of freedom $\nu$: as $\nu \to \infty$, it converges to the Gaussian distribution, while $\nu = 1$ corresponds to the Cauchy distribution.  
We conducted experiments with $\nu \in \{1, 2, 5, \infty\}$, and for each setting, we performed 10 independent trials with different noise realizations.

The results of this numerical experiment are shown in Figure~\ref{fig:S3-non-gaussian}.  
Even when the observation noise followed a non-Gaussian Student’s $t$ distribution, the RRMSE consistently decreased after filtering.  
Under all tested conditions, the post-filtering RRMSE remained below the chance level (RRMSE = 1), indicating effective noise suppression.

These results suggest that the proposed method, based on the EnKF, exhibits robustness to non-Gaussian noise.  
In particular, the Student’s $t$ distribution with $\nu = 1$ (i.e., the Cauchy distribution) has much heavier tails than the Gaussian distribution, allowing extreme outliers to occur more frequently.  
Despite such challenging conditions, the method maintained effective input reconstruction performance.

In the case of Cauchy noise, a reduction in RRMSE was observed even before filtering.  
This effect is primarily attributable to the boundedness of the $\tanh$ activation function used in the ESN, which compresses extreme input values and thereby artificially reduces the reconstruction error.  
This improvement does not reflect enhanced information processing but is rather a trivial effect of activation saturation.

\begin{figure}[H]
\begin{center}
\includegraphics[width=120mm]{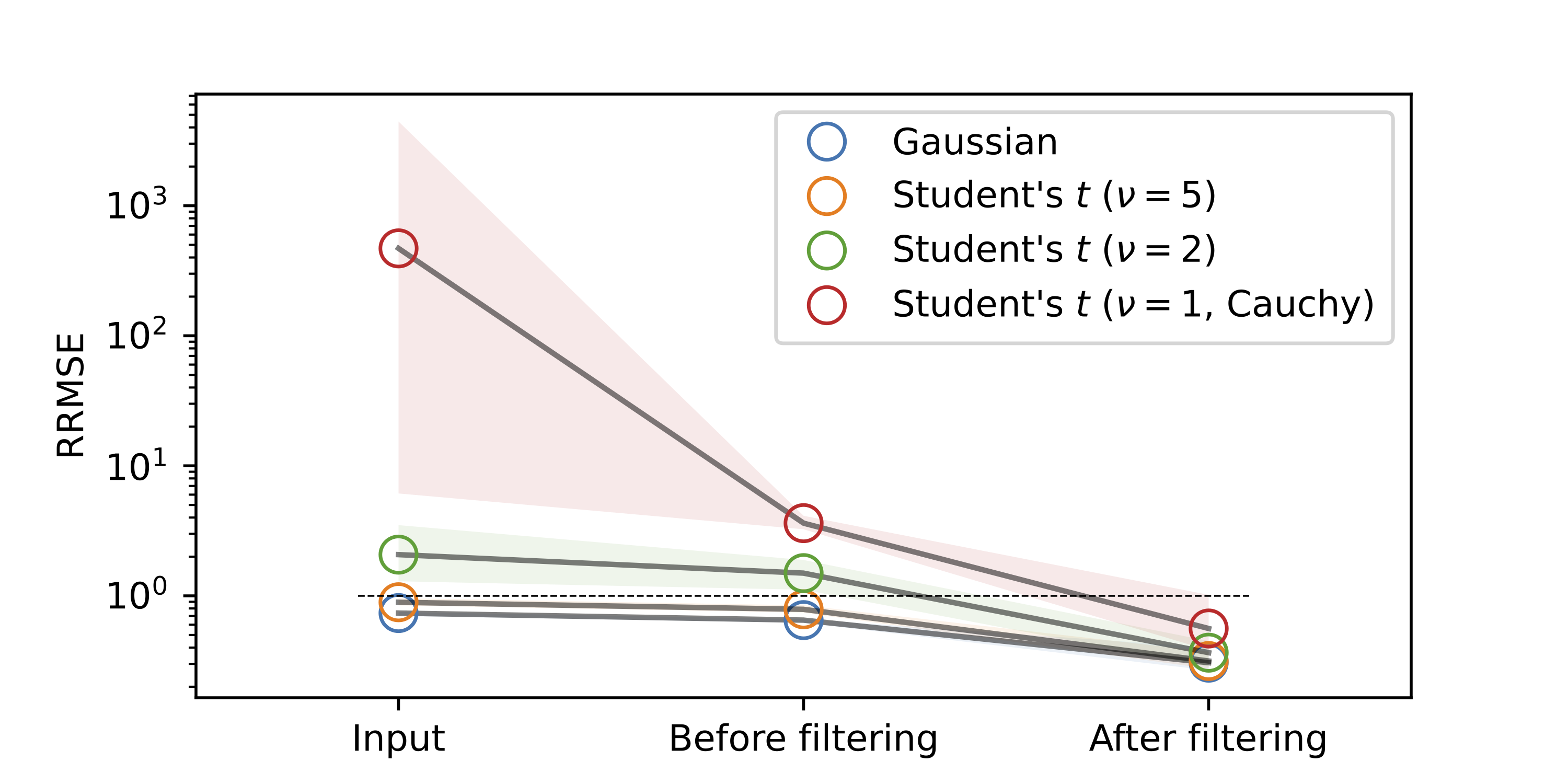}
\caption{
Robustness of input reconstruction to non-Gaussian observation noise.
Root relative mean squared error (RRMSE) was evaluated across three stages: the raw input with noise (“Input”), the reservoir output before noise filtering (“Before filtering”), and the reservoir output after noise filtering (“After filtering”).
Four noise distributions were tested: Gaussian (blue), and Student’s $t$ distributions with degrees of freedom $\nu = 5$ (orange), $\nu = 2$ (green), and $\nu = 1$ (red; Cauchy distribution).
For all distributions, filtering consistently reduced RRMSE.
The black dashed line indicates the chance level (RRMSE = 1).
Shaded areas show the minimum-to-maximum range across 10 trials.
}
\newpage
\label{fig:S3-non-gaussian}
\end{center}
\end{figure}
}

 





\end{document}